\documentclass{article}
\newcommand{\lmn}{FIRST}
\usepackage[preprint]{corl_2026}
\usepackage{enumitem}
\usepackage{graphicx}
\usepackage{wrapfig}
\usepackage{booktabs}
\usepackage{multirow}
\usepackage{amsmath}
\usepackage{tabularx}
\usepackage{amssymb}
\usepackage{algorithm}
\usepackage{algpseudocode}
\usepackage{makecell}
\usepackage{caption}

\usepackage{array}

\title{FACTR 2: Learning External Force Sensing for Commodity Robot Arms Improves Policy Learning}

\author{ \textbf{Steven Oh$^{\ast1,2}$\qquad Jason Jingzhou Liu$^{\ast\dagger1}$\qquad Tony Tao$^{\ast1}$} \\ \textbf{Philip Han$^{1}$ \qquad Kenneth Shaw$^{1}$ \qquad Satoshi Funabashi$^{2}$} \\ \textbf{Ruslan Salakhutdinov$^{1}$ \qquad Deepak Pathak$^{1}$}\\[1mm] $^{1}$Carnegie Mellon University \qquad $^{2}$Waseda University \\ [1mm] $^{\ast}$ Equal contribution}

\begin{document}
\maketitle

\begingroup \renewcommand{\thefootnote}{\fnsymbol{footnote}} \footnotetext[2]{Correspondence to: Jason Jingzhou Liu, \texttt{liujason@cmu.edu}} \endgroup

\vspace{-8mm}
\begin{abstract}
Contact-rich manipulation requires force sensitivity, but many robot arms lack dedicated force sensors due to their high cost. We present \textbf{Neural External Torque Estimation (NEXT)}, a data-driven method that estimates external joint torques without needing any dedicated force sensors. NEXT trains in 1 minute from only 10 minutes of free-motion data, yet achieves estimates comparable to dedicated joint-torque sensors. NEXT enables force-feedback teleoperation on low-cost arms and improves policy learning through \textbf{Force-Informed Re-Sampling Training (FIRST)}, which up-samples pre-contact and contact segments during behavior cloning. Across five long-horizon tasks, FIRST outperforms prior force-aware policies by over 17\% in task progress. Together, NEXT and FIRST bring force-aware teleoperation and policy learning to off-the-shelf robots without additional sensing hardware. Video results and code are available at \url{https://jasonjzliu.com/factr2}

\end{abstract}
\vspace{-2mm}
\begin{figure}[h]
    \centering
    \includegraphics[width=1\linewidth]{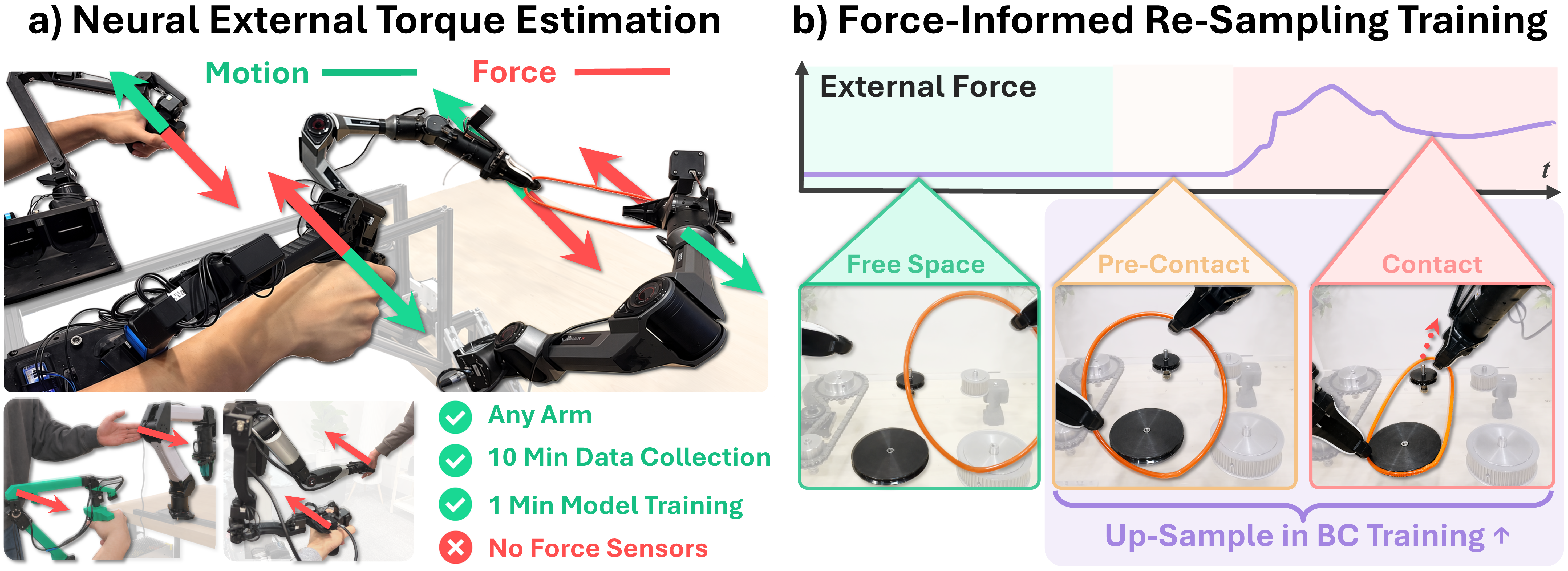}
    \caption{
    \textbf{Overview of our approach.}
    (a) Neural External Torque estimation (NEXT) produces high quality joint torque estimates using only 10 minutes of data without dedicated force sensors or explicit system-identification, enabling force-feedback teleoperation on low-cost arms, such as the Piper, YAM, and Nero.
    (b) Force-Informed Re-Sampling Training (FIRST) uses learned external torque estimates to segment demonstrations into free-space, pre-contact, and contact phases, then up-samples contact-relevant segments during training to improve policy performance.
    }
    \label{fig:placeholder}
\end{figure}

\enlargethispage{1\baselineskip}

\vspace{-5mm}
\section{Introduction}
\vspace{-3mm}
Many everyday manipulation tasks, such as insertion, fastening, and deformable-object handling, require precise force-aware control that humans naturally rely on. In robotics, these subtle contact interactions are difficult to infer reliably from vision alone, making force sensing an essential capability. Dedicated force-torque (FT) sensors are typically available only on expensive platforms such as Franka, Flexiv, or KUKA arms. Many low-cost robot arms lack built-in force sensing, limiting their ability to provide force feedback during data collection and preventing policies from leveraging the force information needed for robust, contact-aware manipulation. Retrofitting platforms with end-effector force sensors remains challenging and expensive, sometimes costing more than the arms themselves. Conventional piezoresistive FT sensors mounted between the arm and end effector rely on multiple strain gauges attached to precision-machined structures, making them expensive and bulky \cite{ATIIndustrialAutomation2026, BotaSystems2024}. Recent capacitive sensors such as CoinFT \cite{choi2025coinft} and magnetic tactile sensors such as ReSkin \cite{bhirangi2021reskin} offer lower-cost alternatives, but their soft sensing materials are not designed to withstand the larger force magnitudes often experienced at the end effector \cite{cao2021six}.

Instead, we present \textbf{N}eural \textbf{Ex}ternal \textbf{T}orque estimation (NEXT), a data-driven method for estimating external joint torque, producing high-quality estimates even on low-cost robot arms. NEXT requires only 10 minutes of contact-free trajectories and 1 minute of training to learn a neural network that predicts the torque required for free-space motion. At deployment, NEXT estimates external torque as the residual between measured motor torque, inferred from motor current, and the predicted free-space torque. Importantly, NEXT does not require dedicated force-torque sensors during either data collection or deployment. We show that NEXT produces signals that closely match dedicated joint-torque sensor measurements on high-end robot arms, while outperforming prior estimation methods.

The learned external force signal enables two capabilities: (1) Force-feedback teleoperation on non-sensorized, low-cost arms (e.g., AgileX Piper, YAM), achieving performance comparable to sensorized systems \cite{liu2025factr, oh2026softwristanisotropicselectable}; and (2) \textbf{F}orce \textbf{I}nformed \textbf{R}e-\textbf{S}ampling \textbf{T}raining (\lmn{}), a behavior cloning method that uses external force estimates to segment demonstrations into free-space, pre-contact, and contact phases, and up-samples the latter two to improve policy success. 

\lmn{} is motivated by the observation that policy failures concentrate not in free-space motion, but in the brief pre-contact intervals requiring precise alignment as well as in contact phases requiring controlled interaction and fine corrections. We evaluate \lmn{} on five long-horizon, contact-rich tasks, where it consistently improves success rates over prior baselines, with gains in alignment, recovery, and contact-phase performance.

With NEXT and FIRST, we call our combined system FACTR 2. It generalizes FACTR~\cite{liu2025factr} beyond arms with dedicated force sensors, enabling both force-feedback teleoperation and force-aware policy learning on commodity robot arms. We believe FACTR 2 democratizes force sensing for any robot arm as we evaluate this capability on platforms spanning low-cost arms such as the Piper (\$2,500) and high-end arms such as the Franka (\$30,000). We summarize our contributions:
\vspace{-2mm}
\begin{itemize}
    \item \textbf{Data-driven external torque estimation without force sensors.} We present NEXT, a neural method for estimating external joint torque without dedicated FT sensors or robot-specific system identification. NEXT trains a recurrent model on only 10 minutes of free-motion data and produces higher-quality estimates than prior methods.

    \item \textbf{Force-feedback teleoperation on low-cost arms.} We use the estimated force signal to enable bilateral teleoperation on non-sensorized low-cost arms, including Piper and YAM, achieving feedback quality comparable to sensorized systems.

    \item \textbf{Force-Informed Re-Sampling Training.} We propose FIRST, a behavior cloning method that uses external torque estimates to identify pre-contact and contact phases in demonstrations, then up-samples these phases during training to improve policy performance.
\end{itemize}

\enlargethispage{\baselineskip}

\vspace{-4mm}
\section{Related Works}
\label{sec:related_work}
\vspace{-2mm}

\subsection{Sensorless External Force Estimation}
\vspace{-2mm}
Estimating external forces without dedicated sensors usually involves modeling robot dynamics and interpreting the residuals between predicted and observed signals as external contact. Traditional approaches based on analytical modeling and system identification \cite{biact, yamane2026, shi2026minimalistcompliancecontrol} can be effective, but they require substantial manual engineering effort and often struggle with actuator nonlinearities like dead zones \cite{inami2024lossfunctionconsideringdead} and torque ripple \cite{zhu2025cycloidal}.  Learning-based methods have emerged to address these limitations. Supervised approaches learn from paired proprioceptive and real world \cite{supervised_anomalydetection} or simulation \cite{liang2021contact} contact data , but require dedicated data collection and labeling. Unsupervised methods such as autoencoder-based anomaly detection \cite{unsupervised_anomalydetection} can identify deviations from manipulator free motion, but are primarily suited for contact detection rather than continuous force estimation. Inverse-dynamics learning has been explored for both individual actuators \cite{zhu2025cycloidal} and complex high-end platforms such as the \textit{da Vinci} system \cite{yilmaz2020neural}, but has not been demonstrated broadly on low-cost robots.

\vspace{-2mm}
\subsection{Imitation Learning with Force Information}
\vspace{-2mm}
Recent imitation learning methods incorporate force or torque information in several ways. One line of work uses force-related signals as an additional policy input. For example, TA-VLA~\cite{tavla} uses joint effort information, BiACT~\cite{biact} uses joint torque, and CompACT~\cite{kamijo2024learning} conditions the policy on end-effector force. ForceVLA~\cite{yu2025forcevla} further introduces a mixture-of-experts architecture for context-aware routing across modality-specific experts. Some methods, such as TA-VLA~\cite{tavla} and FoAR~\cite{he2024force}, use auxiliary objectives to encourage contact-aware representations. A second line of work combines behavior cloning with hybrid force-position or impedance control at inference time~\cite{liu2025forcemimic, filic, xue2025reactive}. These methods predict both motion and force targets, which are then tracked by a low-level controller to enforce the desired interaction behavior. Finally, FACTR~\cite{liu2025factr} uses force information during training to improve policy generalization to unseen objects. Our method, FIRST, uses external torque not only as policy information, but to infer contact phases and use them to reshape the training distribution. Our ablations further show that pre-contact samples are often more valuable than in-contact samples, providing an empirical guideline for force-aware data reweighting.

\vspace{-2mm}
\subsection{Policy Learning with Dataset Reweighting}
\vspace{-2mm}
Policy training dynamics is strongly shaped by how data is used during optimization. In reinforcement learning, prior work has used value and advantage-based weighting to emphasize actions that produce higher returns. AWR \cite{peng2019advantage} weights actions by exponentiated advantages, CRR \cite{wang2020critic} uses a learned critic to filter or weight actions, and AWAC \cite{nair2020awac} applies advantage-weighted updates to bridge offline data and online fine-tuning. In imitation learning, data reweighting has also been explored. Re-Mix adapts DoReMi-style \cite{xie2023doremi} Group DRO for robotics \cite{hejna2025remix}, while DataMIL uses data model estimators \cite{ilyas2022datamodels} for target-aware selection \cite{dass2025datamil}. DemInf scores state-action mutual information to favor diverse, predictable actions \cite{hejna2025mutualifo}, and SCIZOR combines progress prediction and embedding clustering to prune transitions \cite{zhang2025scizor}. Other methods leverage learned value functions for filtering: GVL uses vision-language models for value estimation \cite{ma2024vision}, and SARM trains a stage-aware reward model \cite{chen2025sarm}.  Finally, rollout-based methods optimize closed-loop performance directly. Demo-SCORE filters data using a success classifier trained on rollouts \cite{chen2025demoscore}, and CUPID uses influence functions to estimate how individual demonstrations affect expected returns \cite{agia2025cupid}.

\enlargethispage{1\baselineskip}
\vspace{-3mm}
\section{Background}
\label{sec:background}
\vspace{-3mm}

External joint torque $\tau_{\mathrm{ext}}$ is the joint-space torque induced by physical interactions with the environment, excluding the torque required to move the robot in free space. This signal is directly available on manipulators with dedicated force-torque sensing, but many arms do not have those sensors.

In the absence of dedicated force sensing, external joint torque can be estimated from motor current measurements and a model of the robot dynamics. Motor current is readily provided on most robot arms, and is approximately related to motor torque through $\tau_m = K I_m$, where $I_m$ is the measured motor current and $K$ is the torque constant. Given measured motor torque $\tau_m$ and a known robot dynamics model, the external joint torque can be taken as
\begin{equation}
    \tau_{\mathrm{ext}} = \tau_m - \tau_f,
    \qquad
    \tau_f
    =
    \mathbf{M}(\mathbf{q}) \ddot{\mathbf{q}}
    + \mathbf{C}(\mathbf{q}, \dot{\mathbf{q}}) \dot{\mathbf{q}}
    + \mathbf{g}(\mathbf{q})
    + \tau_{\mathrm{other}}
\end{equation}
Here, $\tau_f$ denotes the torque required to realize the observed motion in free space, i.e., in the absence of external contact. Computing $\tau_f$ therefore requires an accurate model of the robot mass matrix $\mathbf{M}$, Coriolis and centrifugal terms $\mathbf{C}$, gravity term $\mathbf{g}$, and any other contributing terms  $\tau_{\mathrm{other}}$.

However, model-based external torque estimation is highly sensitive to errors in estimating $\tau_f$. In practice, accurately identifying the inertial, gravitational, and friction parameters of a robot is difficult, especially when only motor-side measurements are available \cite{jubien2014dynamic,haddadin2017robot}. Moreover, the residual $\tau_m - \tau_f$ captures not only contact-induced torque, but also unmodeled actuator and transmission effects, including nonlinear friction, stiction, backlash, hysteresis, temperature-dependent drive behavior, sensing noise, torque ripple, deadzones, and saturation \cite{linderoth2013robotic,kircanski1997experimental,reuss2022end}. Because many of these effects are history-dependent and only partially observable from instantaneous joint state, free-space modeling errors can limit purely model-based sensorless force estimation.

\section{Learning External Force Estimation}
\label{sec:learning_sensorless_external_force_estimate}
\vspace{-3mm}

\begin{wrapfigure}{r}{0.52\textwidth}
    \vspace{-3mm}
    \centering
    \includegraphics[width=\linewidth]{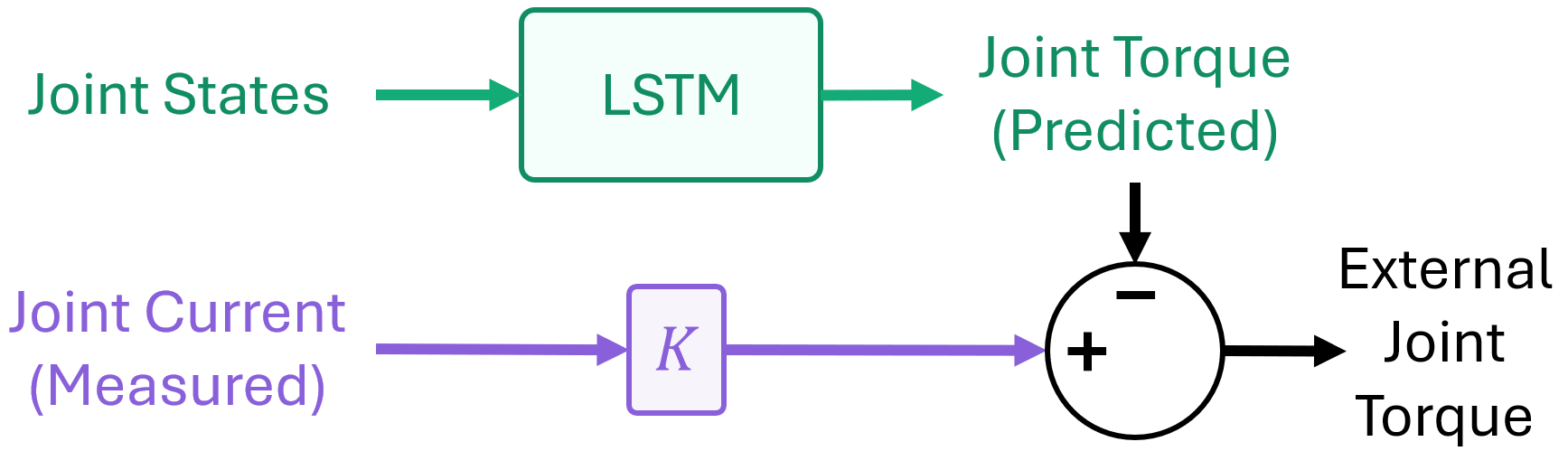}
    \caption{
    \textbf{External Force Estimation Deployment.} At deployment time, we first obtain the measured joint torque from multiplying each joint's measured current by its torque constant $K$. We then use an LSTM trained on free-space data to estimate free-space torque, which is then subtracted from the measured joint torque to obtain external joint torque.
    }
    \label{fig:fsid}
    \vspace{-3mm}
\end{wrapfigure}

We present NEXT, \textbf{N}eural \textbf{Ex}ternal \textbf{T}orque Estimation, a data-driven method for estimating external joint torque $\tau_{\mathrm{ext}}$. Importantly, NEXT does not require dedicated force-torque sensors for neither data collection nor deployment. Instead of computing free-space torque $\tau_f$ from model-based inverse dynamics, we learn a free-space inverse dynamics neural network directly from joint states, predicting the motor torque that would be observed in the absence of external contact:
\begin{equation}
    \hat{\tau}_{f} = f_{\theta}(\mathbf{x}), \quad
    \hat{\tau}_{\mathrm{ext}} = \tau_{m} - \hat{\tau}_{f}.
\end{equation}
where $f_{\theta}$ is a neural network parameterized by $\theta$, and $\mathbf{x}$ is a history of proprioceptive observations. At deployment time, we estimate external joint torque $\hat{\tau}_{\mathrm{ext}}$ by subtracting the predicted free-space torque $\hat{\tau}_{f}$ from the measured motor torque $\tau_m = K I_m$, illustrated in Figure~\ref{fig:fsid}. As shown in Section \ref{sec:next_eval} and Appendix \ref{sec:next_model_ablations}, NEXT yields higher quality external torque estimates against prior methods.

To train the free-space inverse-dynamics model, we collect a dataset of robot arm trajectories executed without external contact,
$
    \mathcal{D}_{\mathrm{free}}
    =
    \{(\mathbf{x}_i, \tau_{m,i})\}_{i=1}^{N},
$
where $\tau_{m,i}$ is the measured motor torque at timestep $i$. The input $\mathbf{x}_i$ is constructed from a temporal history of proprioceptive quantities,
\begin{equation}
    \mathbf{x}_i =
    \left[
    \mathbf{q}_{i-H:i},
    \dot{\mathbf{q}}_{i-H:i},
    \Delta \mathbf{q}_{d,i-H:i}
    \right],
\end{equation}
where $H$ is the history length, $\mathbf{q}$ denotes joint position, $\dot{\mathbf{q}}$ denotes joint velocity, and
$\Delta \mathbf{q}_{d,i}$
is the difference between the commanded joint position target $\mathbf{q}_{d,i}$ and the current joint position $\mathbf{q}_i$.

Because all trajectories in $\mathcal{D}_{\mathrm{free}}$ are collected without external contact, the measured motor torque equals the free-space torque. Thus, we use $\tau_{m,i}$ as the supervision target for learning $\tau_f$. In practice, the dataset can be collected from approximately 10 minutes of free-space robot motion, generated either by teleoperation, motion planning, or replay of known free-space trajectories. Thus, this procedure is easy to adopt on new robot platforms, as it does not require force-torque sensing, payload calibration, or manual system identification.

We instantiate $f_{\theta}$ as an LSTM-based sequence model, which maps the proprioceptive history $\mathbf{x}_i$ to a predicted free-space torque $\hat{\tau}_{f,i}$. The use of temporal history allows the model to capture effects that are difficult to represent with an instantaneous rigid-body dynamics model, such as hysteresis, actuator dynamics, frictional memory, and controller-dependent tracking behavior. The model is trained with an $L_2$ regression loss,
\begin{equation}
    \mathcal{L}(\theta)
    =
    \frac{1}{N}
    \sum_{i=1}^{N}
    \left\|
    f_{\theta}(\mathbf{x}_i) - \tau_{m,i}
    \right\|_2^2
\end{equation}

After training, the model predicts the free-space torque $\hat{\tau}_f$, allowing external torque to be estimated as the residual $\tau_m - \hat{\tau}_f$. Training is lightweight, requiring only 1 minute on an NVIDIA RTX 3090. Details on the model architecture and hyperparameters are provided in Appendix~\ref{sec:next_details}.

\begin{figure}[t]
    \centering
    \includegraphics[width=1\linewidth]{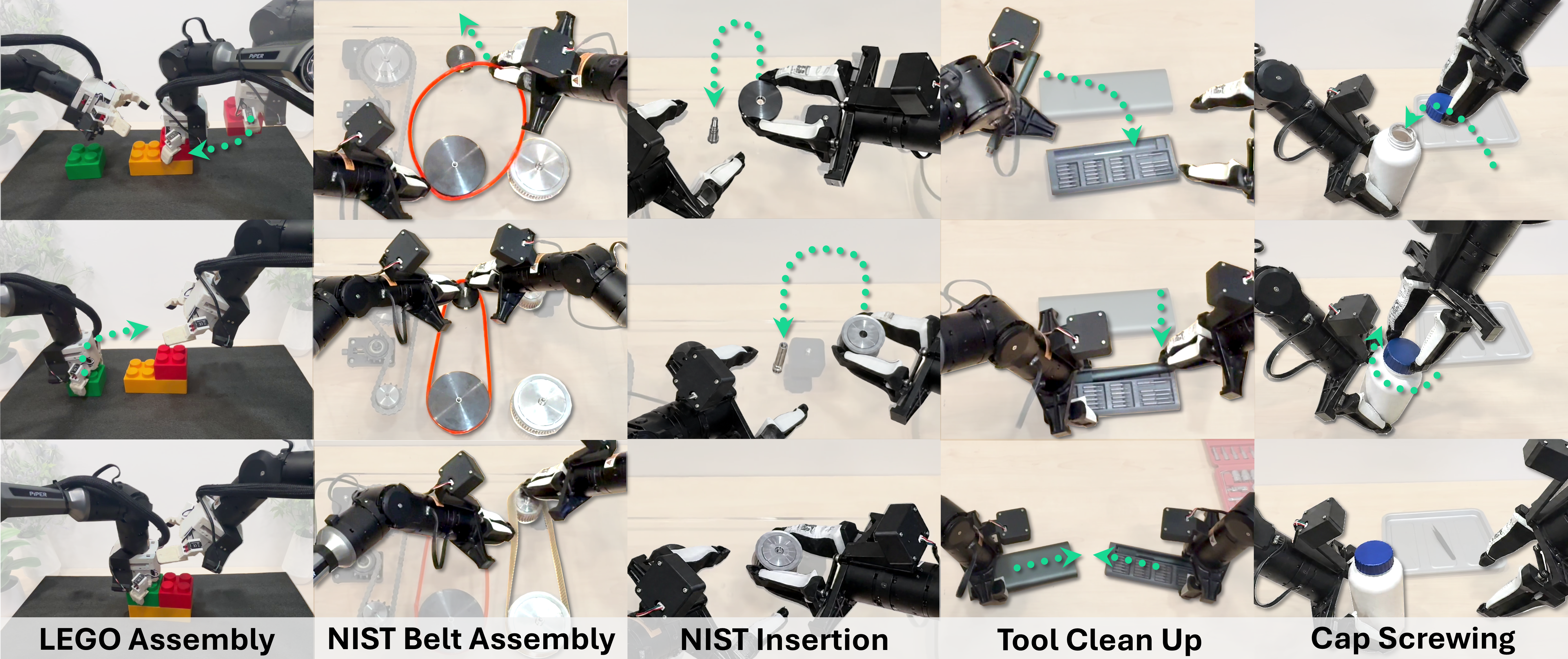}
    \vspace{-5mm}
    \caption{FIRST is evaluated on five long-horizon, contact-rich tasks. Each task comprises multiple stages, many requiring precise alignment or fine-grained, force-sensitive adjustments. Please see videos of these tasks at \url{https://jasonjzliu.com/factr2}
    }
    \label{fig:tasks}
    \vspace{-3mm}
\end{figure}

\section{Force-Informed Re-Sampling Training}
\label{sec:first}
\vspace{-3mm}
To leverage external joint torque for improved behavior cloning, we introduce FIRST, \textbf{F}orce-\textbf{I}nformed \textbf{R}e-\textbf{S}ampling \textbf{T}raining. FIRST is motivated by the observation that failures in contact-rich tasks often occur near contact: pre-contact phases require precise alignment, while contact phases demand controlled interaction and fine corrective actions. FIRST therefore re-weights the training data mixture to sample pre-contact and contact segments more frequently, focusing learning on the most failure-prone portions of the trajectory. In Section~\ref{sec:first_eval}, we show that FIRST improves task success by over 17\% and reduces validation loss on pre-contact and contact phases compared to standard behavior cloning, explaining its downstream performance gains.

\textbf{Problem Statement.} We consider a policy that, at timestep $t$, conditions on an observation
$\mathbf{o}_t =
(\mathbf{I}^{1:M}_t,
\mathbf{q}_t,
\hat{\tau}_{\mathrm{ext},t})$,
consisting of $M$ camera images, joint position $\mathbf{q}_t$, and the estimated external joint torque $\hat{\tau}_{\mathrm{ext},t}$ from Section~\ref{sec:learning_sensorless_external_force_estimate}. The policy predicts a chunk of future joint position targets,
$\hat{\mathbf{a}}_{t:t+k} \sim \pi_{\theta}(\cdot \mid \mathbf{o}_t)$,
where $\mathbf{a}_{t:t+k}$ denotes a corresponding action chunk.

We train the policy by behavior cloning on expert demonstrations
$
    \mathcal{D}
    =
    \left\{
    \left(
    \mathbf{o}_t,
    \mathbf{a}_{t:t+k}
    \right)
    \right\}.
$
We instantiate $\pi_{\theta}$ as a flow-matching policy parameterized by a conditional velocity field $v_{\theta}$. Let $\mathbf{x}_1=\mathbf{a}_{t:t+k}$ be an expert action chunk, $\mathbf{x}_0\sim\mathcal{N}(\mathbf{0},\mathbf{I})$ be noise, and $\mathbf{x}_{\alpha}=(1-\alpha)\mathbf{x}_0+\alpha\mathbf{x}_1$ for $\alpha\sim\mathcal{U}(0,1)$. The policy is trained with a velocity prediction objective \cite{liu2023flow}, where $v_{\theta}$ predicts the flow from noise $\mathbf{x}_0$ to the expert action chunk $\mathbf{x}_1$:
\begin{equation}
    \mathcal{L}_{\mathrm{FM}}(\theta)
    =
    \mathbb{E}
    \left[
    \left\|
    v_{\theta}(\mathbf{x}_{\alpha}, \alpha, \mathbf{o}_t)
    -
    (\mathbf{x}_1 - \mathbf{x}_0)
    \right\|_2^2
    \right].
\end{equation}

\textbf{Segmenting Trajectories into Contact Phases.} To begin, FIRST labels each datapoint in the demonstration dataset as free motion, pre-contact, or contact. Contact corresponds to timesteps where the robot is physically interacting with the environment, pre-contact corresponds to the one-second window before contact onset, and all remaining timesteps are labeled as free motion.

We identify contact from the estimated external joint torque 
$\hat{\tau}_{\mathrm{ext},t}$ using the scalar contact score
$f_t=\|\hat{\tau}_{\mathrm{ext},t}\|_1$. To avoid noisy transitions, we use
hysteresis thresholding: the binary contact indicator $c_t$ turns on when
$f_t \ge T_{\mathrm{high}}$, turns off when $f_t \le T_{\mathrm{low}}$, and
otherwise retains its previous value. 

Contact onsets are timesteps where the label $c_t $ switches from non-contact to contact,
\begin{equation}
    \mathcal{T}_{\mathrm{onset}}
    =
    \left\{
    t \mid c_{t-1}=0,\; c_t=1
    \right\}
\end{equation}
Given a control frequency $F$, we assign the phase label $s_t$ by marking contact timesteps as \texttt{Contact}, the $F$ timesteps before each contact onset as \texttt{Pre-Contact}, and all other timesteps as \texttt{Free Space}:
\begin{equation}
    s_t =
    \begin{cases}
        \texttt{Contact (C)}, 
        & c_t = 1, \\
        \texttt{Pre-Contact (PC)}, 
        & \exists\, t_{\mathrm{onset}}\in\mathcal{T}_{\mathrm{onset}} :
        t_{\mathrm{onset}} - F \leq t < t_{\mathrm{onset}}, \\
        \texttt{Free Space (F)}, 
        & \mathrm{otherwise}.
    \end{cases}
\end{equation}

\textbf{Up-Sampling Pre-Contact and Contact Phases in Training.}
After assigning each datapoint a phase label
$s_t \in \{\texttt{Free Space}, \texttt{Pre-Contact}, \texttt{Contact}\}$,
FIRST modifies only the sampling distribution used for batch construction.
We assign phase-dependent sampling weights $w_{\mathrm{F}}$, $w_{\mathrm{PC}}$,
and $w_{\mathrm{C}}$ to free-space, pre-contact, and contact datapoints,
respectively, with $w_{\mathrm{PC}}, w_{\mathrm{C}} > w_{\mathrm{F}}$.
During training, datapoints are sampled with probability
$p_t = w(s_t)/\sum_{j=1}^{N} w(s_j)$, where $w(s_t)$ denotes the weight
associated with the phase label of datapoint $t$. This weighted random
sampling increases the fraction of pre-contact and contact examples in each
minibatch, emphasizing trajectory segments where precise alignment and
force-sensitive corrections are most critical.

\section{Experiments}
\vspace{-3mm}
\subsection{External Force Estimation and Force-Feedback Teleoperation}
\label{sec:robot_setup}
\vspace{-3mm}
We evaluate external torque estimation on a Franka arm in both contact and free-space settings. We use the Franka because it has built-in joint torque sensors and a factory-calibrated external torque estimate, providing a high-quality reference which we use as ground truth during contact.
For contact, we collect 5 minutes of held-out data, unseen during training, while a human applies external forces to the robot. We report the $L_1$ error averaged across joints between the ground-truth sensor signal and estimates from NEXT (ours), FILIC~\cite{filic}, and a disturbance observer (DO)~\cite{mamedov2020practical}. For free-space motion, where the true external torque should be zero, we collect 5 minutes of held-out data without contact and use zero torque as ground truth, treating the Franka sensor signal as another baseline. Baseline details are provided in Appendix~\ref{sec:NEXT_baselines_impl}.  We repeat the free-space evaluation on the AgileX Piper, a low-cost arm without force sensors, again using zero torque as ground truth. Since the Piper lacks external torque sensing, we cannot repeat the contact  evaluation; instead, we assess contact performance through a force-feedback teleoperation user study.

\begin{wrapfigure}{r}{0.49\textwidth}
    \centering
    \vspace{-4mm}
    \includegraphics[width=\linewidth]{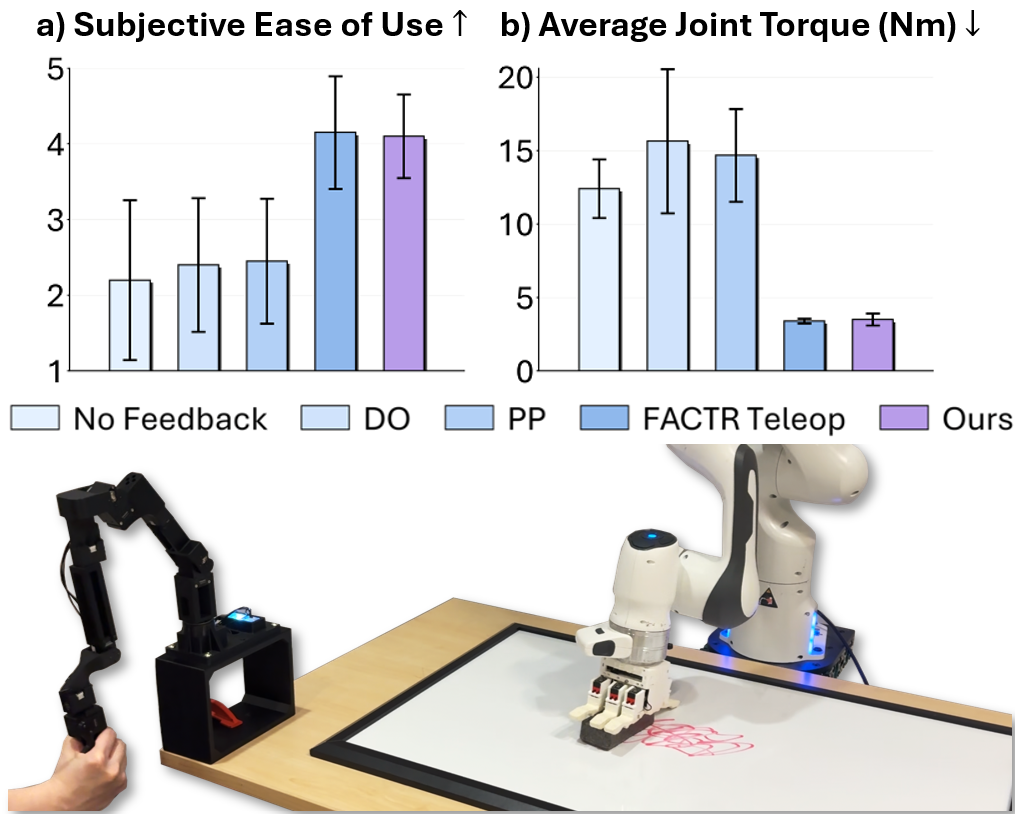}
    \caption{\textbf{User Study.} Our external force estimate (using NEXT) improves teleoperation over baselines and remains comparable to FACTR Teleop, which relies on dedicated external force sensors.  Lower joint torque applied corresponds to less unnecessary exertion.}
    \label{fig:user_study}
    \vspace{-3mm}
\end{wrapfigure}

\textbf{Force-Feedback Teleoperation User Study.}
We evaluate whether the learned external torque estimate provides useful teleoperation feedback. 
In a user study with 20 participants, users perform a wiping task on the Franka arm, as shown in Figure~\ref{fig:user_study}, 
with 5 feedback conditions: no feedback, disturbance-observer based feedback (DO), leader-follower position-position feedback \cite{bilateral_survey} (PP), FACTR Teleop using Franka’s built-in torque sensors, and FACTR Teleop with NEXT's external torque estimates. 
Note that both FACTR Teleop variants implement the following force-feedback control law \cite{liu2025factr}:
\vspace{-1mm}
\begin{equation}
    \mathbf{\tau}_{feedback} = \mathbf{K}_{f,p} \mathbf{\tau}_{ext}
\end{equation}
where $\mathbf{\tau}_{feedback}$ is joint torque applied to the leader arm, and $\mathbf{K}_{f,p}$ scales the external joint torque $\mathbf{\tau}_{ext}$ sensed by the follower arm. In this study, $\mathbf{\tau}_{ext}$ either comes from Franka's factory external torque estimate or comes from NEXT. We collect 1--5 ease-of-use ratings and measure the applied joint torque. We repeat the study on the Piper arm. Details about the baselines are described in Appendix \ref{sec:teleoperation_baselines_impl}.
\enlargethispage{\baselineskip}

\vspace{-2mm}
\subsection{Imitation Learning with Force Input}
\label{sec:bc_setup}
\vspace{-2mm}

\textbf{Experiment Settings.} 
We evaluate FIRST against baselines on five long-horizon, contact-rich manipulation tasks, shown in Figure~\ref{fig:tasks}. Detailed task descriptions are provided in Appendix~\ref{sec:task_description}. For each task, we collect 250 demonstrations on a bimanual Piper setup with four RGB cameras: two wrist cameras and two overhead cameras. Each policy is evaluated over 20 rollouts per task. We report task progress, with scoring criteria defined in Appendix~\ref{sec:task_progress_rate}.

\textbf{Policy Details and Baselines.} 
We compare FIRST with four baselines while keeping the policy architecture and behavior cloning objective fixed. All methods use a flow-matching policy trained for 20 epochs, with an architecture adapted from Multi-modal Diffusion Transformer~\cite{reuss2024multimodal} and an action chunk size of 30. We also repeat the experiments with Action Chunking Transformer (ACT)~\cite{zhao2023learning}. 
The action head of each policy is randomly initialized and trained from scratch. Implementation details for both are provided in Appendix~\ref{sec:policy_details}. We compare the following variants:
\vspace{-1mm}
\begin{itemize}[leftmargin=1em,itemsep=0pt,topsep=0pt,parsep=0pt,partopsep=0pt]
    \item \textit{Base Policy:} Policy takes in images and robot joint positions as inputs, with no force information.
    \item \textit{Base Policy + Torque:} Base policy with the learned external torque estimate as additional input.
    \item \textit{FACTR~\cite{liu2025factr}:} Uses external torque and blurs input images during training with a noise schedule.
    \item \textit{TA-VLA~\cite{tavla}:} Uses an auxiliary objective to reconstruct the input external torque.
    \item \textit{FIRST (Ours):} Uses learned external torque as input and up-samples pre-contact and contact data during training. We ablate up-sampling ratios in Appendix~\ref{sec:upsampling_ablations}.
\end{itemize}

\begin{figure}[!t]
\centering

\begin{minipage}[t]{0.43\textwidth}
\vspace{0pt}
\centering
\scriptsize
\setlength{\tabcolsep}{3pt}
\renewcommand{\arraystretch}{1.05}

\resizebox{\linewidth}{!}{%
\begin{tabular}{llc}
\toprule
\textbf{Motion Type} & \textbf{External Torque Method} & \textbf{$L_1$ Error (Nm)} \\
\midrule
\multirow{4}{*}{Contact}
& External Sensor (GT) & $0.000 \pm 0.000$ \\
& FILIC~\cite{filic} & $4.395 \pm 1.531$ \\
& Disturbance Observer~\cite{mamedov2020practical} & $1.471 \pm 0.761$ \\
& \textbf{NEXT (Ours)} & $\mathbf{0.547 \pm 0.348}$ \\
\midrule
\multirow{5}{*}{Free Space}
& Free Space (GT) & $0.000 \pm 0.000$ \\
& FILIC~\cite{filic} & $2.460 \pm 1.358$ \\
& Disturbance Observer~\cite{mamedov2020practical} & $2.429 \pm 1.298$ \\
& External Sensor & $0.449 \pm 0.208$ \\
& \textbf{NEXT (Ours)} & $\mathbf{0.414 \pm 0.278}$ \\
\bottomrule
\end{tabular}%
}

\captionof{table}{
Average $L_1$ joint torque error on the Franka in free-space and contact settings. Free-space errors are measured against zero external torque, while contact errors are measured against the Franka built-in external torque estimate. NEXT achieves the lowest error in both settings.
}
\label{tab:avg_joint_error}
\end{minipage}
\hfill
\begin{minipage}[t]{0.54\textwidth}
\vspace{0pt}
\centering
\includegraphics[width=1\linewidth]{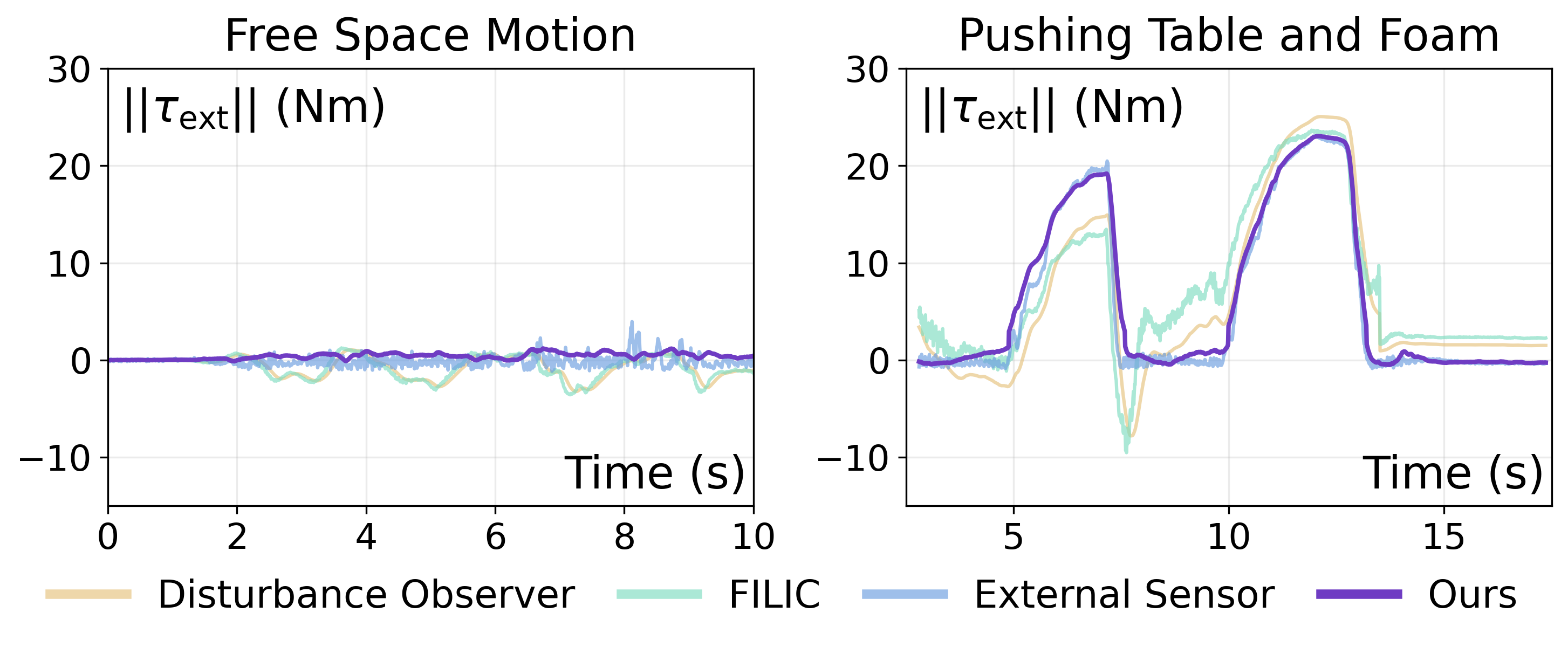}
\captionof{figure}{
\textbf{Left:} In free space, the external joint torque should remain zero. Our method produces an estimate that is less noisy than the external sensor and remains near zero, while FILIC and the Disturbance Observer drift away.
\textbf{Right:} During contact, our estimate closely tracks the external sensor, whereas FILIC and the Disturbance Observer deviate substantially.
}
\label{fig:force_plots_comparisons}
\end{minipage}
\vspace{-5mm}
\end{figure}

\section{Results}
\vspace{-3mm}
\label{sec:result}
In our external force estimation experiments, we ask: (i) How do NEXT’s external joint torque estimates compare to ground-truth sensor measurements and baselines? (ii) How effective is NEXT for force-feedback teleoperation compared to alternative feedback methods?

In our imitation learning experiments, we ask: (i) How does FIRST compare to other baselines leveraging force? (ii) Which contact phases should FIRST up-sample for best performance?

\textbf{NEXT predicts external joint torque comparable to dedicated sensors.}
\label{sec:next_eval} We present quantitative results for NEXT on the Franka system in Table~\ref{tab:avg_joint_error}. Under external contact, NEXT closely matches the Franka external joint torque sensor, achieving an average $L_1$ error of only $0.547$ Nm, which is $87.6\%$ lower than FILIC and $62.8\%$ lower than the disturbance observer. This trend is also shown qualitatively on the right side of Figure~\ref{fig:force_plots_comparisons}: NEXT tracks the sensor signal closely, while the baselines deviate substantially, especially during contact transitions.

In free space, where the true external torque should be zero, we treat the Franka's external torque sensor signals as an additional baseline; NEXT achieves the lowest error, even outperforming this dedicated sensor. As shown on the left side of Figure~\ref{fig:force_plots_comparisons}, its estimate is less noisy and remains closer to zero than the sensor. This can be attributed to NEXT learning hardware-specific effects directly from data collected on the hardware itself, yielding high-quality force estimates.

We observe similar results on the AgileX Piper arm, where NEXT achieves an average joint $L_1$ error of only $0.018$ Nm, substantially lower than the baselines; full results are provided in Table~\ref{tab:piper_force_results} in Appendix~\ref{sec:piper_evaluations}. We also ablate the NEXT model architecture in Appendix~\ref{sec:next_model_ablations}, showing that the LSTM outperforms both GRU and MLP variants.

\textbf{NEXT enables high-quality force-feedback teleoperation on low-cost arms.}
In the Franka user study, participants rated NEXT-based force feedback as easier to use than the baselines, shown in Figure~\ref{fig:user_study} (a). NEXT also reduced the force required to complete the task, performing comparably to FACTR Teleop, which uses dedicated joint torque sensors, seen in Figure~\ref{fig:user_study} (b). Lower applied force indicates greater force awareness and less unnecessary exertion. On the Piper arm, which lacks dedicated force sensors, NEXT also outperforms the baselines, as detailed in Appendix~\ref{sec:piper_evaluations}. Overall, these results show that NEXT provides high quality force feedback without needing force sensors, comparable to sensor-based methods.

\begin{figure}[t]
    \centering
    \includegraphics[width=1\linewidth]{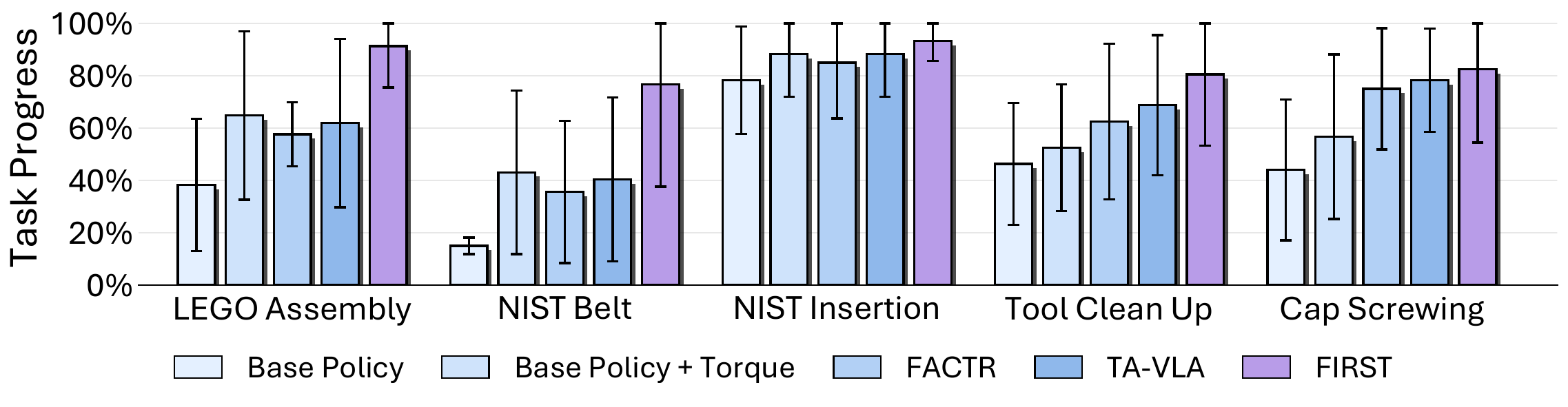}
    \caption{
    Task progress across five contact-rich manipulation tasks using flow matching policy.
    }
    \vspace{-1em}
    \label{fig:task_progress}
\end{figure}

\label{sec:first_eval}
\textbf{FIRST leads to improved policy performance.}
In Figure ~\ref{fig:task_progress}, we present the main qualitative results for FIRST against baselines for the flow matching policy, with ACT results shown in Appendix \ref{sec:act}. FIRST achieves the highest task progress across all five real-world tasks. The improvement is especially clear in long-horizon, multi-stage tasks, such as NIST Belt, where successful execution requires reliable transitions between free motion, pre-contact alignment, and contact-rich manipulation. Qualitatively, we observe that FIRST produces more robust recovery behaviors before key contact events, such as fine-grained realignment motions before insertion, wrapping, or screwing.

\begin{wrapfigure}{r}{0.5\textwidth}
    \centering
    \includegraphics[width=\linewidth]{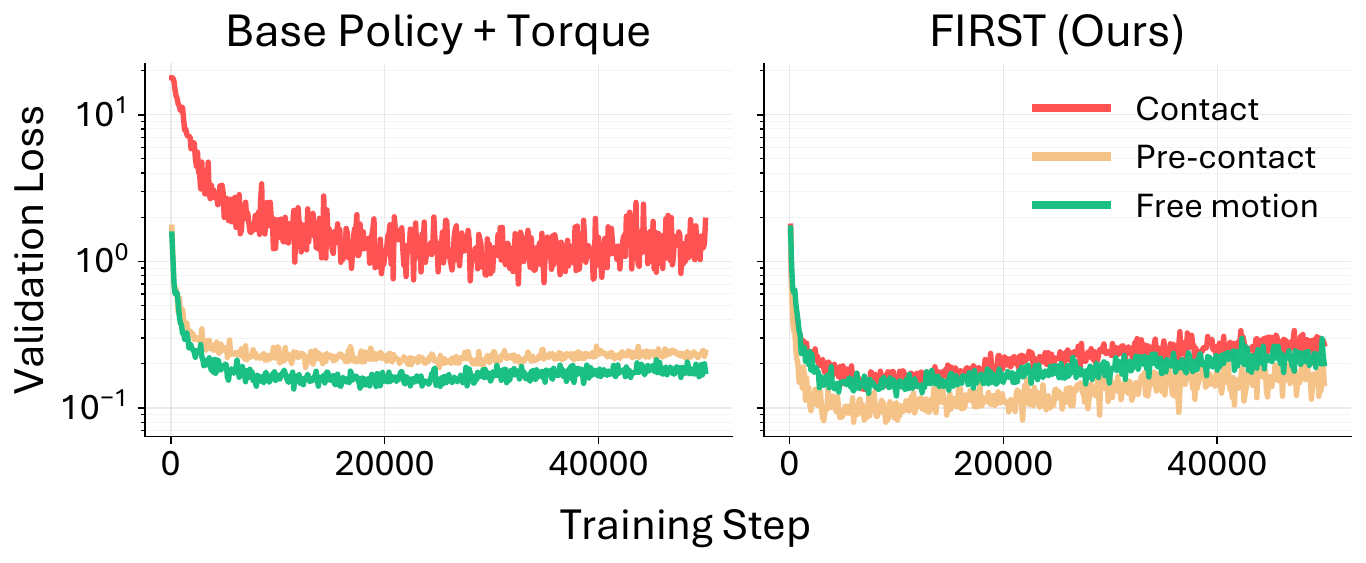}
    \vspace{-1.5em}
    \caption{
     Default sampling yields higher validation loss on pre-contact and contact phases. By upsampling these phases during training, FIRST reduces their validation losses.
    }
    \label{fig:val_loss}
    \vspace{-1.5em}
\end{wrapfigure}

Figure~\ref{fig:val_loss} further explains this improvement through phase-wise validation loss for the NIST Belt task. For the Base Policy + Torque baseline, loss is higher on pre-contact and contact samples than on free-motion samples, indicating that these phases are harder to model. By upsampling pre-contact and contact data, FIRST reduces their validation losses and brings them closer to the free-motion loss, focusing training on the phases most critical for contact-rich manipulation.

\textbf{FIRST upsampling ablation.} 
We ablate which phases to upsample for best policy performance. As shown in Table~\ref{tab:upsampling_ablation}, upsampling pre-contact samples achieves the highest average task progress, 

\begin{wraptable}{r}{0.46\textwidth}
\vspace{-1em}
\centering
\small
\setlength{\tabcolsep}{3.5pt}
\renewcommand{\arraystretch}{1.08}

\begin{tabular*}{\linewidth}{@{\extracolsep{\fill}}lccc@{}}
\toprule
& \multicolumn{3}{c}{\textbf{Upsampling Phase}} \\
\cmidrule(lr){2-4}
\textbf{Task} & \textbf{C} & \textbf{PC} & \textbf{PC+C} \\
\midrule
LEGO Assembly  & 0.697 & \textbf{0.913} & 0.875 \\
NIST Belt      & 0.494 & \textbf{0.767} & 0.649 \\
NIST Insertion & 0.800 & \textbf{0.933} & \textbf{0.933} \\
Tool Clean Up  & 0.650 & \textbf{0.805} & 0.775 \\
Cap Screwing   & 0.707 & 0.673 & \textbf{0.825} \\
\midrule
\textbf{Average} & 0.670 & \textbf{0.818} & 0.811 \\
\bottomrule
\end{tabular*}

\caption{
\textbf{Upsampling ablation.}
C: contact, PC: pre-contact, PC+C: pre-contact + contact.
}
\vspace{-5mm}
\label{tab:upsampling_ablation}
\end{wraptable}

\vspace{-2mm}
suggesting that the moments before contact are especially important as they determine reaching the correct pose and alignment. Upsampling both pre-contact and contact phases is also competitive, and performs best on Cap Screwing, where sustained contact is essential. In contrast, upsampling only contact samples does not improve performance. Overall, these results show that emphasizing contact-relevant phases, especially pre-contact, improves long-horizon manipulation. Additional results on the highest performing upsampling ratio for each task are provided in Appendix~\ref{sec:upsampling_ablations}.

\textbf{Why not use raw joint current as input?}
Since low-cost arms often expose motor current, we compare policies conditioned on raw current with those conditioned on NEXT-estimated external torque. As shown in Table~\ref{tab:current_vs_external_torque}, raw current improves NIST Insertion but not Cap Screwing, while NEXT-estimated external torque performs best on both tasks. This suggests that subtracting learned free-space dynamics yields a cleaner contact representation than raw current alone, further highlighting the usefulness of NEXT. Full results and attention analysis are in Appendix~\ref{sec:current_vs_external_torque}.

\enlargethispage{\baselineskip}

\section{Conclusion}
\label{sec:conclusion}
We present NEXT, a data-driven method for estimating external joint torque without dedicated force sensors. With only 10 minutes of free-space data and 1 minute of training, NEXT produces estimates comparable to dedicated sensor measurements and enables force-feedback teleoperation on low-cost arms. We further introduce FIRST, which uses NEXT torque estimates to up-sample pre-contact and contact segments during behavior cloning, improving performance on contact-rich tasks by emphasizing the phases most critical for alignment and interaction.

\newpage
\section{Limitations}
NEXT has two main limitations. First, its absolute torque scale depends on the motor torque constant $K$, which is typically provided by the manufacturer or derived from a current--torque curve. If $K$ is inaccurate, absolute scale requires calibration against an external force sensor. In practice, this is less limiting for force-feedback teleoperation, where forces are often scaled for comfort~\cite{liu2025factr}, and also for policy learning, where inputs are normalized anyways. Second, NEXT is robot-specific and may require retraining across arms due to hardware variation. 

\acknowledgments{
We thank Andrew Wang, Yulong Li, Koki Yamane, Yangcen Liu, Yongliang Wang, Ritvik Singh, Zheyuan Hu, Adam Kan, and Arthur Allshire for valuable discussions and feedback on the paper. We also thank Zheyuan Hu and members of the AIRe Lab for generously providing access to their YAM arms for this project. This work was supported in part by AFOSR FA9550-23-1-0747, ONR MURI N00014-22-1-2773 and ONR MURI N00014-24-1-2748.
}

\bibliography{citation}  
\appendix
\clearpage
\raggedbottom                                     

\section*{Appendix}
\section{NEXT Implementation Details}
\label{sec:next_details}
\subsection{Free-Space Inverse Dynamics Model Architecture}
We use an LSTM-based sequence model for free-space inverse dynamics prediction. Rather than maintaining a persistent hidden state across the entire trajectory, we adopt a stateless sliding-window formulation. The network therefore implicitly models temporal dynamics within the look-back window, while the hidden state is reset for each sequence. This formulation provides recent motion context without accumulating hidden-state drift over long trajectories. The local history allows the model to capture velocity-dependent, actuator-dependent, and unmodeled friction effects without explicitly estimating acceleration. For deployment, the estimator can be evaluated using a rolling buffer of the most recent $N$ observations. External joint torque is then estimated as the residual between the measured joint torque and the predicted free-space torque. The network architecture and training hyperparameters are listed in Table \ref{tab:lstm_hyperparameters}.

On a single thread of an Intel Core i9-10920X CPU, the free-space inverse dynamics model required approximately 1.76 ms per forward pass, corresponding to an effective update rate of 568 Hz. This enables high-frequency force feedback teleoperation. In practice, we run the model at 100 Hz.

\begin{table}[!htbp]
\centering
\small
\renewcommand{\arraystretch}{1.12}
\setlength{\tabcolsep}{6pt}
\begin{tabularx}{\textwidth}{@{}lX@{}}
\toprule
\textbf{Hyperparameter} & \textbf{Value} \\
\midrule
Recurrent backbone & 2-layer LSTM with hidden size 128 \\
Prediction head & 2-layer MLP with hidden size 256 \\
Dropout & 0.1 \\
Loss function & Mean squared error (MSE) \\
Optimizer & AdamW \\
Base learning rate & $1 \times 10^{-3}$ \\
Weight decay & $1 \times 10^{-6}$ \\
Gradient clipping & 1.0 \\
Early stopping & Enabled; validation loss, patience 20, warmup 10, $\min\Delta=10^{-5}$ \\
Hardware & NVIDIA RTX 3090, 24 GB \\
Training time & Approximately 1 minute with early stopping \\
Data frequency & 100Hz \\
\bottomrule
\end{tabularx}
\vspace{1mm}
\caption{
Network architecture and training hyperparameters for the free-space inverse dynamics neural network used in NEXT.
}
\label{tab:lstm_hyperparameters}
\end{table}

\vspace{-5mm}
\subsection{Free-Space Inverse Dynamics Network Ablations}
\label{sec:next_model_ablations}
To select the free-space inverse dynamics model used in NEXT, we ablate model architecture, input modality, and history length using external torque estimation error in both free-space and contact settings. Prior work has used MLPs for torque estimation on specialized \textit{da Vinci} systems~\cite{yilmaz2020neural} and GRUs for single-actuator external torque estimation~\cite{zhu2025cycloidal}. We therefore compare MLP, GRU, and LSTM architectures.

Table~\ref{tab:NEXT_network_ablations} reports the architecture ablation results, with values corresponding to the average joint $L_1$ validation error between predicted free-space torque and ground-truth free-space torque on the held-out set. MLPs perform worse than recurrent models, showing the importance of temporal history for free-space torque estimation. GRUs perform well in free space, but performs worse during contact. LSTMs achieve the best free-space and contact performance, where longer-term actuator, frictional, and controller-dependent effects may play a larger role.

We further ablate input modalities and history length. Specifically, we compare histories of joint positions $q$, joint positions and velocities $(q,\dot{q})$, and joint positions, velocities, and tracking error $(q,\dot{q},\Delta q_d)$, with history lengths of 10, 25, and 50 steps. Using $(q,\dot{q},\Delta q_d)$ consistently outperforms the other inputs, suggesting that tracking error provides useful information about controller effort and actuator response. Short histories of 10 steps generally yield higher error, while increasing the history to 50 steps improves performance, especially for LSTM models. Based on these results, we use an LSTM with history length $H=50$ and input $(q,\dot{q},\Delta q_d)$ as the final NEXT free-space inverse dynamics model.

\begin{table*}[t]
\centering
\scriptsize
\setlength{\tabcolsep}{2.5pt}
\renewcommand{\arraystretch}{1.05}
\resizebox{\textwidth}{!}{%
\begin{tabular}{llcccccc}
\toprule
\multirow{2}{*}{\textbf{Input}} 
& \multirow{2}{*}{$\mathbf{H}$}
& \multicolumn{2}{c}{\textbf{MLP}}
& \multicolumn{2}{c}{\textbf{GRU}}
& \multicolumn{2}{c}{\textbf{LSTM (Ours)}} \\
\cmidrule(lr){3-4}
\cmidrule(lr){5-6}
\cmidrule(lr){7-8}
& 
& \textbf{Free Motion $\downarrow$} & \textbf{Contact $\downarrow$}
& \textbf{Free Motion $\downarrow$} & \textbf{Contact $\downarrow$}
& \textbf{Free Motion $\downarrow$} & \textbf{Contact $\downarrow$} \\
\midrule
\multirow{3}{*}{$q$}
& 10 
& $2.35{\pm}2.34$ & $1.78{\pm}1.64$
& $2.25{\pm}1.79$ & $1.12{\pm}1.09$
& $2.53{\pm}2.52$ & $2.61{\pm}2.29$ \\

& 25 
& $2.02{\pm}2.01$ & $1.34{\pm}1.32$
& $1.68{\pm}1.30$ & $0.665{\pm}0.478$
& $2.21{\pm}2.20$ & $1.45{\pm}1.39$ \\

& 50 
& $1.76{\pm}1.76$ & $1.13{\pm}1.06$
& $1.62{\pm}1.24$ & $0.843{\pm}0.835$
& $2.15{\pm}2.15$ & $1.09{\pm}0.908$ \\

\midrule

\multirow{3}{*}{$q,\dot{q}$}
& 10 
& $1.04{\pm}1.03$ & $0.681{\pm}0.647$
& $0.863{\pm}0.862$ & $0.580{\pm}0.535$
& $1.15{\pm}1.15$ & $0.715{\pm}0.553$ \\

& 25 
& $0.980{\pm}0.976$ & $0.908{\pm}0.802$
& $0.814{\pm}0.814$ & $0.596{\pm}0.506$
& $1.11{\pm}1.11$ & $0.551{\pm}0.511$ \\

& 50 
& $1.02{\pm}1.02$ & $0.905{\pm}0.892$
& $0.795{\pm}0.794$ & $0.653{\pm}0.549$
& $1.12{\pm}1.12$ & $0.578{\pm}0.526$ \\

\midrule

\multirow{3}{*}{$q,\dot{q},\Delta q_d$}
& 10 
& $0.706{\pm}0.702$ & $1.89{\pm}1.71$
& $0.423{\pm}0.422$ & $1.94{\pm}1.75$
& $0.504{\pm}0.503$ & $0.876{\pm}0.244$ \\

& 25 
& $0.755{\pm}0.747$ & $1.56{\pm}1.45$
& $0.473{\pm}0.472$ & $1.47{\pm}1.29$
& $0.490{\pm}0.489$ & $0.789{\pm}0.453$ \\

& 50
& $0.813{\pm}0.811$ & $1.47{\pm}1.33$
& $0.429{\pm}0.429$ & $1.62{\pm}1.45$
& $\mathbf{0.414{\pm}0.278}$ & $\mathbf{0.547{\pm}0.348}$ \\

\bottomrule
\end{tabular}
}
\caption{
Ablation of model family, input modality, and temporal history length for external torque estimation. Values corresponding to the average joint $L_1$ validation error between predicted free-space torque and ground-truth free-space torque on the held-out set, reported in Nm.
}
\label{tab:NEXT_network_ablations}
\end{table*}

\subsection{Free-Space Data Collection Guidelines for NEXT}
\begin{figure}[h]
    \centering
    \includegraphics[width=0.9\linewidth]{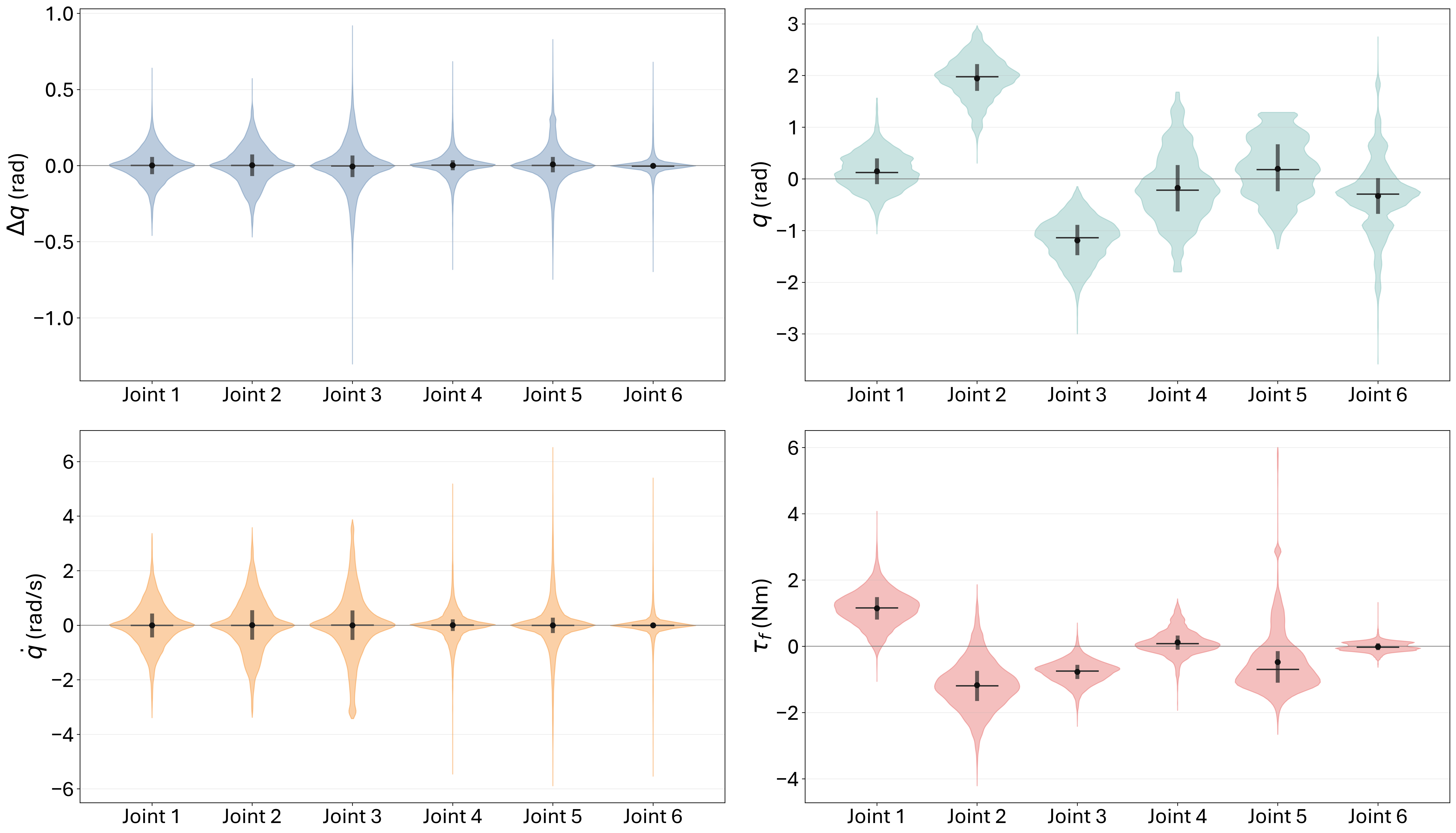}
    \caption{
    Distribution of the 10-minute free-motion dataset used to train NEXT on the Piper arm.
    The plots show the coverage of joint-position error $\Delta q$, joint position $q$, joint velocity $\dot{q}$, and free-space torque $\tau_f$ across all six joints.
    }
    \label{fig:next_data_dis}
\end{figure}

Figure~\ref{fig:next_data_dis} shows the distribution of the 10-minute free-motion dataset collected on the Piper arms for training the free-space inverse dynamics LSTM for NEXT.
During data collection, we follow a simple coverage-oriented procedure: first, each joint is moved independently across its range of motion; next, the end effector is moved through Cartesian-like motions involving multiple joints; finally, both slow and fast motions are repeated to capture velocity-dependent dynamics.
Although these trajectories can be generated automatically through motion planning, in practice they can also be recorded once and replayed on new robot setups to quickly collect training data.

\vspace{-5mm}
\subsection{External Joint Torque Estimation Baseline Implementation}
\label{sec:NEXT_baselines_impl}
We implement two model-based external joint torque estimation baselines for NEXT: FILIC~\cite{filic} and a disturbance observer~\cite{mamedov2020practical}. We evaluate both methods on the Franka Panda and AgileX Piper arms. For the Franka, we use the dynamics model from~\cite{franka_sysid}, whose parameters were obtained through optimization-based system identification. For the AgileX Piper, we use the dynamics model derived from the manufacturer-provided URDF~\cite{agilexrobotics_agx_arm_urdf}.

\textbf{FILIC \cite{filic}.}
For FILIC, we use the standard residual-based estimate introduced in Sec.~\ref{sec:background}. Given the measured motor torque $\tau_m$, external torque is estimated by subtracting the nominal free-space torque computed using MuJoCo inverse dynamics:
\begin{equation}
    \hat{\tau}_{\mathrm{ext}}^{\mathrm{FILIC}}
    =
    \tau_m - \tau_f,
    \qquad
    \tau_f
    =
    \mathbf{M}(\mathbf{q}) \ddot{\mathbf{q}}
    + \mathbf{C}(\mathbf{q}, \dot{\mathbf{q}}) \dot{\mathbf{q}}
    + \mathbf{g}(\mathbf{q})
\end{equation}
This method directly attributes the residual between measured motor torque and modeled free-space dynamics to external interaction. 

\textbf{Disturbance Observer~\cite{mamedov2020practical}.}
As a second baseline, we implement a momentum-based disturbance observer. Instead of directly computing acceleration-dependent inverse dynamics, the observer estimates external torque from the residual in generalized momentum,
\begin{equation}
    \mathbf{p}
    =
    \mathbf{M}(\mathbf{q})\dot{\mathbf{q}} .
\end{equation}
It maintains an internal momentum estimate $\hat{\mathbf{p}}$ with dynamics
\begin{equation}
    \dot{\hat{\mathbf{p}}}
    =
    \tau_m
    - \mathbf{C}(\mathbf{q}, \dot{\mathbf{q}})\dot{\mathbf{q}}
    - \mathbf{g}(\mathbf{q})
    + \dot{\mathbf{M}}(\mathbf{q})\dot{\mathbf{q}}
    + \hat{\tau}_{\mathrm{ext}}^{\mathrm{MDO}},
    \qquad
    \hat{\tau}_{\mathrm{ext}}^{\mathrm{MDO}}
    =
    \mathbf{K}_o(\mathbf{p}-\hat{\mathbf{p}}).
\end{equation}
By avoiding direct computation of $\ddot{\mathbf{q}}$, the momentum observer is less sensitive to encoder noise and produces smoother estimates than FILIC. As shown in Figure~\ref{fig:force_plots_comparisons}, it is less noisy than FILIC but remains less accurate than NEXT.

\subsection{Force-Feedback Teleoperation Baseline Implementation}
\label{sec:teleoperation_baselines_impl}

For the force-feedback teleoperation user study, we implement five feedback conditions on the Franka Panda:

\begin{itemize}[leftmargin=1em,itemsep=0pt,topsep=0pt,parsep=0pt,partopsep=0pt]
    \item \textit{No Feedback:} A GELLO leader-follower system~\cite{wu2023gello} without force feedback.

    \item \textit{Disturbance Observer (DO):} External joint torque is estimated on the follower arm using the disturbance observer described in Appendix~\ref{sec:NEXT_baselines_impl}, then relayed to the leader arm as force feedback.

    \item \textit{Position-Position Feedback (PP):} Feedback torque is computed by scaling the joint position difference between the leader and follower arms, following~\cite{bilateral_survey}.

    \item \textit{FACTR Teleop:} The FACTR teleoperation system~\cite{liu2025factr}, which feeds back external joint torque from dedicated joint torque sensors, along with gravity compensation, friction compensation, and null-space regulation.

    \item \textit{Ours:} The same teleoperation framework as FACTR Teleop, but using NEXT-predicted external joint torque instead of dedicated joint torque sensor measurements.
\end{itemize}

We repeat the same study on the AgileX Piper arm, excluding FACTR Teleop because the Piper does not provide dedicated joint torque sensing.

\begin{figure}[t]
    \centering

    \begin{minipage}[t]{0.50\linewidth}
        \centering
        \vspace{0pt}
        \begin{tabular}{lc}
            \toprule
            External Torque Method & Error (Nm) $\downarrow$ \\
            \midrule
            Free Motion (GT) & $0.000 \pm 0.000$ \\
            FILIC & $0.435 \pm 0.225$ \\
            Disturbance Observer & $0.248 \pm 0.132$ \\
            \textbf{Ours} & $\mathbf{0.018 \pm 0.012}$ \\
            \bottomrule
        \end{tabular}
        \captionof{table}{
        Average $L_1$ joint torque error on the AgileX Piper in free-space settings. 
        Free-space errors are measured against zero external torque. NEXT achieves the lowest error.
        }
        \label{tab:piper_force_results}
    \end{minipage}
    \hfill
    \begin{minipage}[t]{0.46\linewidth}
        \centering
        \vspace{0pt}
        \includegraphics[width=0.7\linewidth]{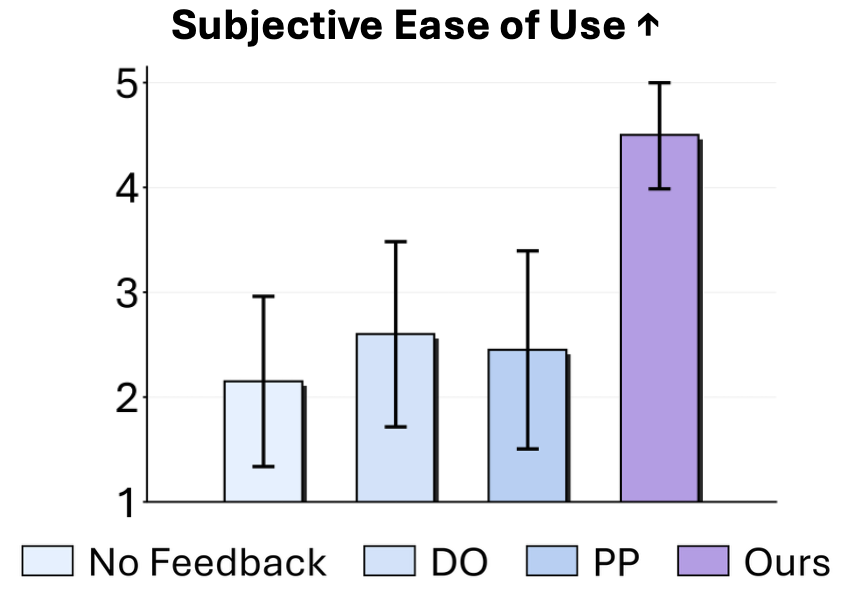}
        \captionof{figure}{
        User study results comparing external torque feedback methods on the Piper.
        }
        \label{fig:piper_user_study}
    \end{minipage}

    \vspace{-3mm}
\end{figure}

\subsection{AgileX Piper NEXT Evaluations}
\label{sec:piper_evaluations}
Since the AgileX Piper does not provide dedicated external joint torque sensing, we evaluate torque estimation only in the free-space setting, where the ground-truth external torque is known to be zero. As shown in Table~\ref{tab:piper_force_results}, NEXT achieves the lowest error, with an average $L_1$ error of only $0.018$ Nm, outperforming both FILIC and the disturbance observer. These results show that NEXT can produce accurate external torque estimates even on a low-cost arm without dedicated force sensors.

To evaluate NEXT in contact settings, we perform a force-feedback teleoperation user study. As shown in Figure~\ref{fig:piper_user_study}, participants rated teleoperation with NEXT-based force feedback as easier to use than the baseline feedback methods.

\section{Behavior Cloning Policy Architecture and Training Hyperparameters}
\label{sec:policy_details}

Our behavior cloning policy takes RGB images, proprioceptive inputs, and force estimates as input. Each modality is tokenized before being passed to either an ACT or Flow Matching DiT policy architecture. Images are encoded with a DINOv3-Base encoder~\cite{simeoni2025dinov3}, while proprioceptive and force inputs are projected into tokens using MLP layers.

Proprioceptive inputs include joint positions and gripper widths, and force inputs correspond to estimated external joint torques. We use only the most recent observation at the current timestep. The action space consists of absolute joint angles and gripper width, with an action prediction horizon of 30. We note that the action head of each policy is randomly initialized and trained from scratch. Training details for both ACT and Flow Matching policies are provided in Table~\ref{tab:policy_hyperparameters}.

\begin{table}[t]
\centering
\renewcommand{\arraystretch}{1.08}
\begin{tabularx}{\textwidth}{@{}p{0.42\textwidth}X@{}}
\toprule
\textbf{Hyperparameter} & \textbf{Value} \\
\midrule

\multicolumn{2}{c}{\textbf{Behavior Policy Training}} \\
\midrule
Optimizer & AdamW \\
Base Learning Rate & $3 \times 10^{-4}$ \\
Weight Decay & $0.05$ \\
Optimizer Momentum & $\beta_1, \beta_2 = 0.9, 0.95$ \\
Batch Size & 128 \\
Learning Rate Schedule & Cosine Decay \\
Total Epochs & 15--20 \\
Warmup Steps & 500 \\
Augmentation & RandomResizeCrop \\
GPU & A6000 (48 GB) \\
Wall-Clock Time & 12--24 hours \\

\midrule
\multicolumn{2}{c}{\textbf{Action Chunking Transformer Details}} \\
\midrule
\# Encoder Layers & 4 \\
\# Decoder Layers & 6 \\
\# MHSA Heads & 8 \\
Hidden Dim & 768 \\
Feed-Forward Dim & 3072 \\
Dropout & 0.1 \\
Positional Encoding & Sinusoidal \\
Action Chunk & 30 \\

\midrule
\multicolumn{2}{c}{\textbf{Flow Matching Details}} \\
\midrule
\# Decoder Layers & 6 \\
\# MHSA Heads & 8 \\
Hidden Dim & 768 \\
Feed-Forward Dim & 3072 \\
Dropout & 0.1 \\
Positional Encoding & Sinusoidal \\
Action Chunk & 30 \\
Eval Steps & 10 \\
Time Sampling & $\tau \sim \mathrm{Beta}(1.0, 1.5)$ \\
ODE Solver & Euler \\
Noise Distribution & $\mathcal{N}(0,I)$ \\
\bottomrule
\end{tabularx}
\vspace{1mm}
\caption{Policy Architecture and Training Hyperparameters}
\label{tab:policy_hyperparameters}
\end{table}

\section{Behavior Cloning Evaluation Tasks}
\subsection{Task Description}
\label{sec:task_description}
We evaluate FIRST along with baselines on five long-horizon, contact-rich manipulation tasks:
\begin{itemize}[leftmargin=1em,itemsep=0pt,topsep=0pt,parsep=0pt,partopsep=0pt]
    \item \textit{LEGO assembly (LEAP Hand V2 \cite{shaw2024leap2}):} bimanual hand assembly of two small LEGO blocks onto a bigger LEGO block.
    \item \textit{NIST belt assembly (gripper):} bimanual gripper manipulation of two belts around two pulley sets.
    \item \textit{NIST insertion (gripper):} insertion of two pulley sets onto pegs on a NIST board.
    \item \textit{Tool Clean Up (gripper):} screw driver insertion followed by case assembly.
    \item \textit{Cap screwing (gripper):} cap pickup, alignment, and screwing onto a container until tightened.
\end{itemize}

\subsection{Task Progress Rate Calculation}
\label{sec:task_progress_rate}

We report task progress rate in addition to binary success rate to evaluate partial completion in multi-stage manipulation tasks. Each task is divided into several stages, and each stage is marked as either completed (\texttt{s}) or failed (\texttt{f}) for every rollout. The task progress rate is calculated as the percentage of completed stages across all rollouts.


The task stages are listed in Table~\ref{tab:task_progress_stages}. For example, the LEGO task consists of grasping, aligning, and inserting both the red and green cubes. The NIST Belt task includes picking up and aligning both belt components before pushing them into place. The NIST Insertion task measures progress through picking, aligning, and inserting the black part, followed by picking up the aluminum part. The Case Assembly task evaluates whether the robot can pick up, align, and insert the driver, then assemble the box.

For the cap screwing task, we use a modified progress score because a policy can fail by continuing to rotate the cap after it is already tightened. Therefore, we subtract a penalty based on the number of over-turns. Each over-turn reduces the progress score by \(0.02\). This penalizes policies that complete the screwing motion but do not stop properly.

\begin{table}[t]
\centering
\setlength{\tabcolsep}{5pt}
\renewcommand{\arraystretch}{1.1}
\begin{tabular}{ll}
\toprule
\textbf{Task} & \textbf{Stages} \\
\midrule
LEGO 
& red grasp, red align, red insert, green grasp, green align, green insert \\

NIST Belt 
& pick orange, align big, align small, push, pick grey, align grey, push \\

NIST Insertion 
& pick black, align black, insert black, pick aluminum \\

Tool Clean Up 
& pick driver, align driver, insert driver, assemble box \\

Cap Screwing 
& pick cap, align cap, screw on, end properly \\
\bottomrule
\end{tabular}
\vspace{1mm}
\caption{Stage decomposition used for task progress calculation.}
\label{tab:task_progress_stages}
\end{table}

\subsection{Imitation Learning with Force Information Baseline Implementations}
\label{sec:policy_baseline_details}

\textbf{Base Policy:} We use the base ACT or flow-matching policy described in Sec.~\ref{sec:bc_setup}. The policy takes multi-view RGB images and robot proprioception, including joint positions and gripper states, as input and predicts a chunk of future joint and gripper commands.

\textbf{Base Policy + Torque:} This baseline augments the base policy with external joint torque predictions from NEXT. The predicted torque signal is embedded with an MLP and added to the policy context as an additional observation token.

\textbf{FACTR~\cite{liu2025factr}:} FACTR keeps the policy architecture unchanged and applies a visual curriculum to all image observations. Following the original FACTR implementation, we use a Gaussian blur operator with a linear curriculum schedule. Specifically, at training step $n$, the blur strength is set as
\[
\sigma_n = \sigma_0 \left(1 - \frac{n}{N}\right),
\]
where $\sigma_0$ is the initial blur strength and $N$ is the total number of curriculum steps. Thus, the input images are strongly blurred at the beginning of training, and the blur is gradually removed as training progresses.

\textbf{TA-VLA\cite{tavla}:} TA-VLA uses the predicted torque/contact signal as an auxiliary learning target in addition to the action. During inference, only the robot-action dimensions are executed, while the auxiliary torque/contact prediction is used only to shape the learned representation during training.

\section{Additional Results on FIRST}

\subsection{Segmenting Trajectories into Contact Phases Visualization}
\begin{figure}[h]
    \centering
    \includegraphics[width=1\linewidth]{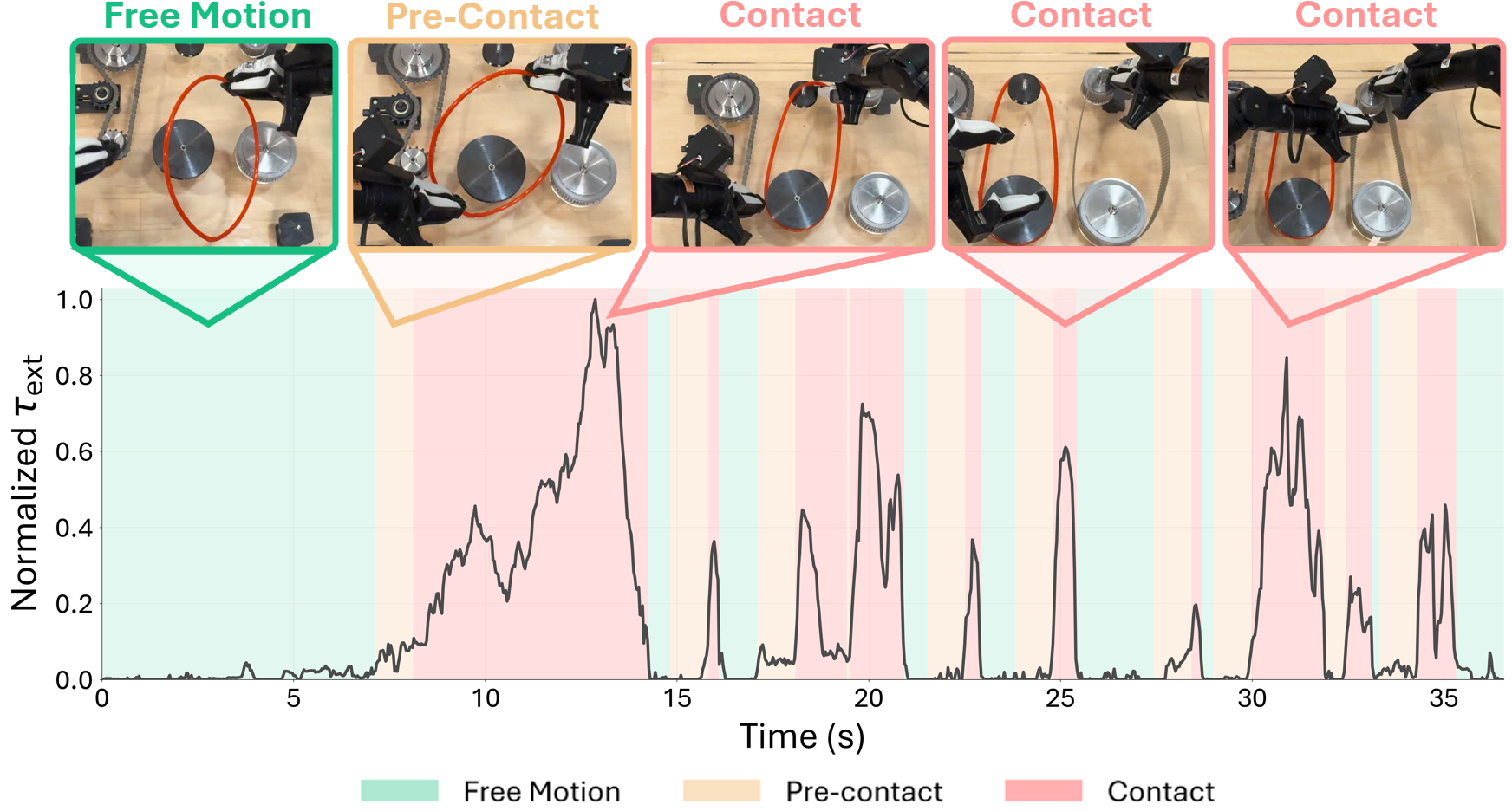}
    \caption{
    \textbf{Contact phase segmentation from learned external torque.}
    The normalized $\tau_{ext}$ from NEXT is used to segment demonstrations into free motion, pre-contact, and contact phases. The image overlays show that the predicted labels align with the robot's interaction state during the task.
    }
    \label{fig:auto_labeling}
\end{figure}

Figure ~\ref{fig:auto_labeling} shows an example of phase labels produced by our hysteresis-threshold-based labeling procedure. The estimated external torque from NEXT remains near zero during free motion and rises clearly during physical interaction, allowing the trajectory to be separated into free-motion, pre-contact, and contact phases. The pre-contact label captures the short interval immediately before contact onset, where precise alignment is required, while the contact label captures periods of sustained interaction. This sensor-level signal enables automatic segmentation of demonstrations even on robot arms without dedicated force sensors, providing the phase annotations used by FIRST for force-informed re-sampling.

\subsection{Task Progress Evaluation on ACT}
\label{sec:act}
\begin{figure}[h]
    \centering
    \includegraphics[width=1\linewidth]{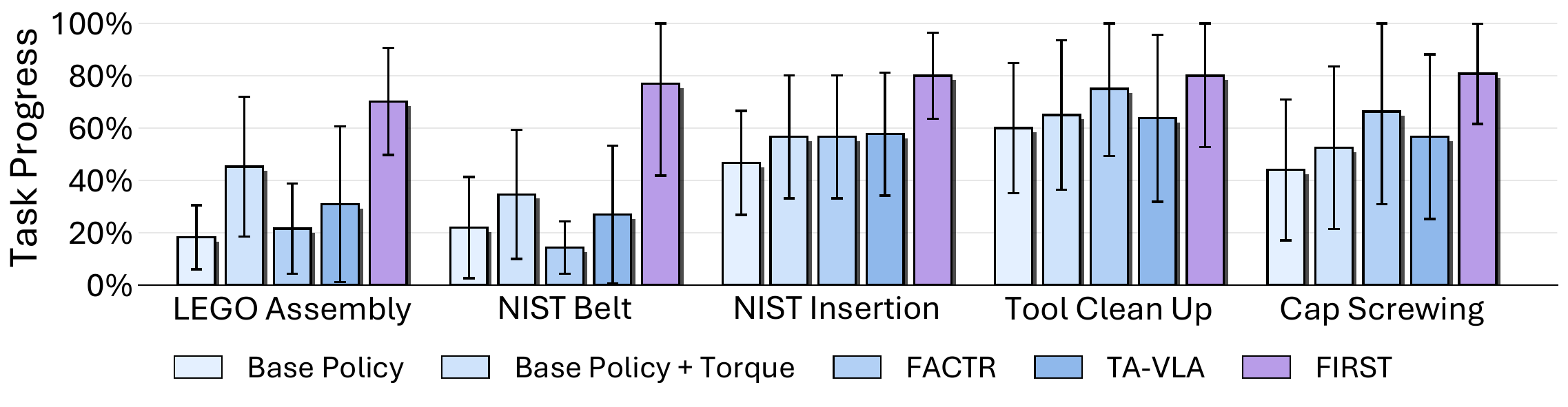}
    \caption{Task progress across five contact-rich manipulation tasks using ACT policy.}
    \label{fig:act}
\end{figure}

We also apply FIRST to ACT, in addition to the flow-matching policy presented in Section~\ref{sec:first_eval}. As shown in Figure~\ref{fig:act}, we observe a similar trend: upsampling pre-contact and contact phases improves task progress over the baselines.

\subsection{Selected Re-Sampling Ratios for Each Task}
\label{app:best_resampling_ratios}

We report the task-specific re-sampling ratios used by FIRST in Table~\ref{tab:best_resampling_ratios}. 
All ratios are written as \(w_{\mathrm{F}} : w_{\mathrm{PC}} : w_{\mathrm{C}}\), where \(w_{\mathrm{F}}\), \(w_{\mathrm{PC}}\), and \(w_{\mathrm{C}}\) denote the sampling weights for free-motion, pre-contact, and contact samples, respectively. 
For most tasks, FIRST uses \(1:5:1\), which up-samples pre-contact samples by \(5\times\) relative to free-motion and contact samples. 
For Cap Screwing, FIRST uses \(1:3:3\), up-sampling both pre-contact and contact samples, since sustained contact is central to successful execution. 
These ratios are selected based on task progress for each task.

\begin{table}[h]
\centering
\small
\begin{tabular}{lc}
\toprule
\textbf{Task} & \textbf{Best Re-Sampling Ratio} \\
\midrule
LEGO Assembly & \(1:5:1\) \\
NIST Belt Assembly & \(1:5:1\) \\
NIST Insertion & \(1:5:1\) \\
Tool Clean Up & \(1:5:1\) \\
Cap Screwing & \(1:3:3\) \\
\bottomrule
\end{tabular}
\vspace{1mm}
\caption{Best re-sampling ratios used by FIRST for each task. Ratios are reported as \(w_{\mathrm{F}} : w_{\mathrm{PC}} : w_{\mathrm{C}}\), corresponding to free-motion, pre-contact, and contact sampling weights.}
\label{tab:best_resampling_ratios}
\vspace{-5mm}
\end{table}

\subsection{Up-Sampling Magnitude Ablation}
\label{sec:upsampling_ablations}

\begin{figure}[h]
    \centering
    \includegraphics[width=1\linewidth]{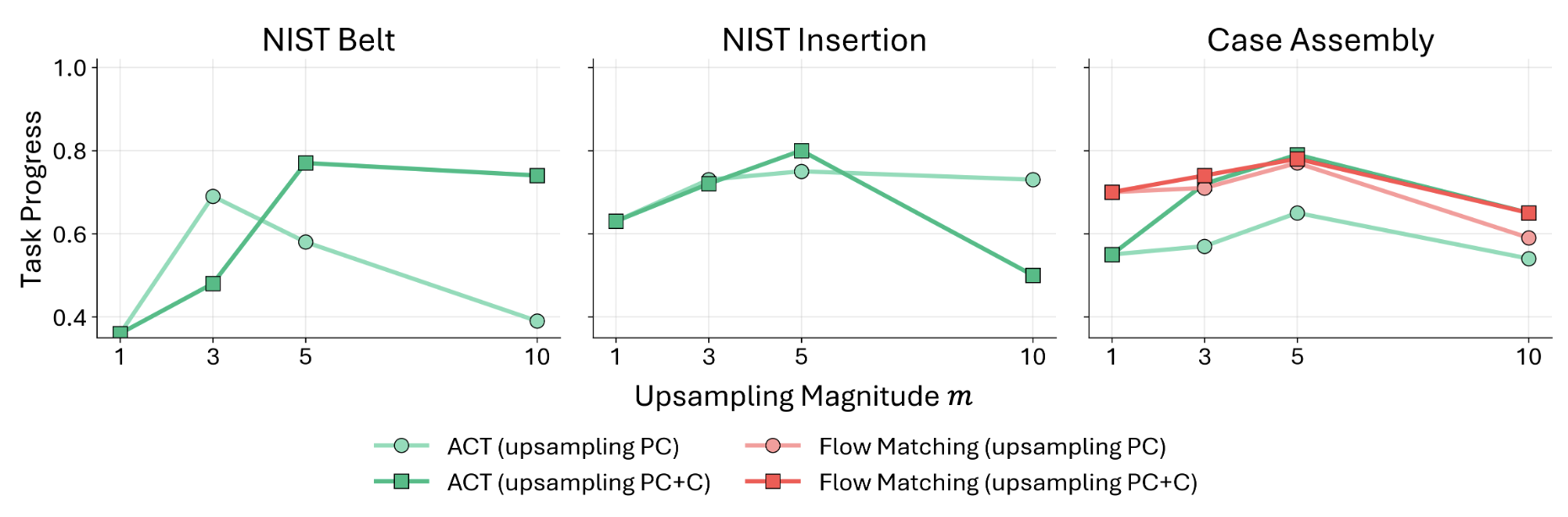}
    \caption{
    Moderate up-sampling generally improves performance, with best results around \(m=5\), while larger values lead to saturation or degradation.
    }
    \label{fig:upsampling_mag}
\end{figure}

Figure~\ref{fig:upsampling_mag} ablates the effect of up-sampling magnitude on task progress. 
We define the magnitude \(m\) as the sampling weight assigned to the up-sampled phase relative to free motion. 
For pre-contact up-sampling, the phase weights are
\[
w_{\mathrm{F}} : w_{\mathrm{PC}} : w_{\mathrm{C}} = 1 : m : 1,
\]
whereas for pre-contact + contact up-sampling, the phase weights are
\[
w_{\mathrm{F}} : w_{\mathrm{PC}} : w_{\mathrm{C}} = 1 : m : m.
\]
We run this ablation on NIST Belt Assembly, NIST Insertion, and Tool Clean Up using ACT. 
Due to limited time and compute, we additionally run the flow-matching ablation only on Tool Clean Up. 
Across tasks, moderate up-sampling improves task progress, with performance generally peaking around \(m=5\). 
Larger values lead to saturation or degradation, suggesting that overly aggressive up-sampling can over-bias the training distribution toward contact-relevant segments and reduce coverage of the full task trajectory.

\subsection{Up-Sampling Policies Without Force Input}
\label{sec:upsampling_without_force_input}

While Section~\ref{sec:first_eval} evaluates FIRST on policies that condition on external joint torque, FIRST can also be applied without using force as a policy input. 
In this setting, NEXT is used only offline to segment demonstrations into free-motion, pre-contact, and contact phases. 
These phase labels are then used by FIRST to re-sample the training data, increasing the frequency of pre-contact and contact examples in each minibatch. 
Importantly, the policy itself receives only the standard visual and proprioceptive observations, without the predicted external joint torque.

Table~\ref{tab:upsampling_with_without_force} highlights two trends. 
First, when comparing methods within the columns without external joint torque, phase-based upsampling improves over standard behavior cloning. 
For both ACT and Flow Matching, using PC or PC+C upsampling generally increases task success compared to no upsampling, showing that NEXT-derived phase labels are useful even when force estimates are used only at training time for data re-sampling. 
This indicates that FIRST can improve standard behavior cloning by reshaping the training distribution toward task-critical segments where alignment and contact-sensitive corrections are most important.

Second, when comparing horizontally between policies trained with and without external joint torque, the force-conditioned policies consistently achieve higher success rates. 
This shows that while phase-based upsampling alone provides a clear benefit, directly feeding the predicted external joint torque into the policy provides additional state information that further improves contact-rich manipulation. 
Overall, the best performance is achieved when both components are used together: NEXT provides external joint torque as a policy input, while FIRST uses NEXT-derived phase labels to up-sample the most relevant pre-contact and contact phases during training.

\begin{table}[!htbp]
\centering
\small
\setlength{\tabcolsep}{3.5pt}
\renewcommand{\arraystretch}{1.12}
\begin{tabular}{llcccc}
\toprule
\multirow{2}{*}{\textbf{Policy}} 
& \multirow{2}{*}{\makecell{\textbf{Up-sampling}\\\textbf{Strategy}}}
& \multicolumn{2}{c}{\makecell{\textbf{w/o $\tau_{ext}$}}} 
& \multicolumn{2}{c}{\makecell{\textbf{w/ $\tau_{ext}$}}} \\
\cmidrule(lr){3-4} 
\cmidrule(lr){5-6}
& 
& \makecell{\textbf{Belt}\\\textbf{Assembly}} 
& \makecell{\textbf{Tool}\\\textbf{Clean Up}} 
& \makecell{\textbf{Belt}\\\textbf{Assembly}} 
& \makecell{\textbf{Tool}\\\textbf{Clean Up}} \\
\midrule

\multirow{3}{*}{ACT}
& None    & 0.220 & 0.600 & 0.347 & 0.650 \\
& PC      & 0.360 & 0.600 & 0.687 & 0.650 \\
& PC + C  & 0.283 & 0.763 & 0.770 & 0.788 \\

\midrule
\multirow{3}{*}{Flow Matching}
& None    & 0.150 & 0.463 & 0.431 & 0.525 \\
& PC      & 0.213 & 0.575 & 0.767 & 0.805 \\
& PC + C  & 0.241 & 0.525 & 0.649 & 0.775 \\
\bottomrule
\end{tabular}
\vspace{1mm}
\begin{flushleft}
\footnotesize
\end{flushleft}
\caption{
Effect of phase-based upsampling on policies trained with and without external joint torque.
Results are reported as task success rates.
}
\label{tab:upsampling_with_without_force}
\end{table}

\subsection{Inputting Joint Current vs. External Joint Torque on Policy Performance}
\label{sec:current_vs_external_torque}

We further examine whether raw joint current is sufficient for contact-aware policy learning, or whether the learned external joint torque estimate provides a more useful representation. 
Prior work such as TA-VLA~\cite{tavla} uses readily available joint current measurements from low-cost robot arms as a force-related input for imitation learning. 
However, joint current is only an indirect measure of physical interaction: it contains actuator effort required for both free-space motion and external contact, as well as hardware- and controller-dependent effects. 
As a result, the policy must implicitly disentangle contact-induced torque from free-space dynamics.

In contrast, NEXT explicitly estimates external joint torque by subtracting a learned free-space torque prediction from the measured motor torque. 
This distinction is important because estimating a clean external torque signal from current alone generally requires temporal context. 
As shown in our NEXT free-space inverse dynamics ablations in Appendix \ref{sec:next_model_ablations}, longer proprioceptive histories improve torque estimation, suggesting that instantaneous current and state observations are insufficient to reliably separate free-space actuation from contact. 
A standard behavior cloning policy, however, typically receives only the current observation rather than a long proprioceptive history such as the 50-step history used by NEXT. 
While one could provide such a history directly to the policy, this increases the burden on the policy to simultaneously infer dynamics, localize contact, and solve the visuomotor task. 
In practice, long proprioceptive histories may also encourage overfitting to trajectory-specific motion patterns rather than using vision appropriately to react to scene variation. 
NEXT avoids this issue by using history in a specialized force-estimation module, while providing the policy with a compact, contact-relevant external torque signal.

We compare policies conditioned on raw joint current and learned external joint torque on two contact-sensitive tasks: NIST Insertion and Cap Screwing. 
Both tasks require force cues for successful execution: NIST Insertion depends on detecting alignment and contact during insertion, while Cap Screwing requires identifying when the cap has been fully tightened. 
As shown in Table~\ref{tab:current_vs_external_torque}, conditioning on joint current improves performance on NIST Insertion compared to the base ACT policy, but does not improve Cap Screwing. 
In contrast, conditioning on learned external joint torque achieves the best performance on both tasks. 
Qualitatively, we observe that policies conditioned on raw joint current often continue rotating the cap after it is already tightened, leading to over-screwing failures. 
Policies conditioned on learned external torque are better able to detect the tightened contact state and terminate the screwing motion appropriately.

\begin{figure}[!htbp]
    \centering
    \includegraphics[width=0.65\linewidth]{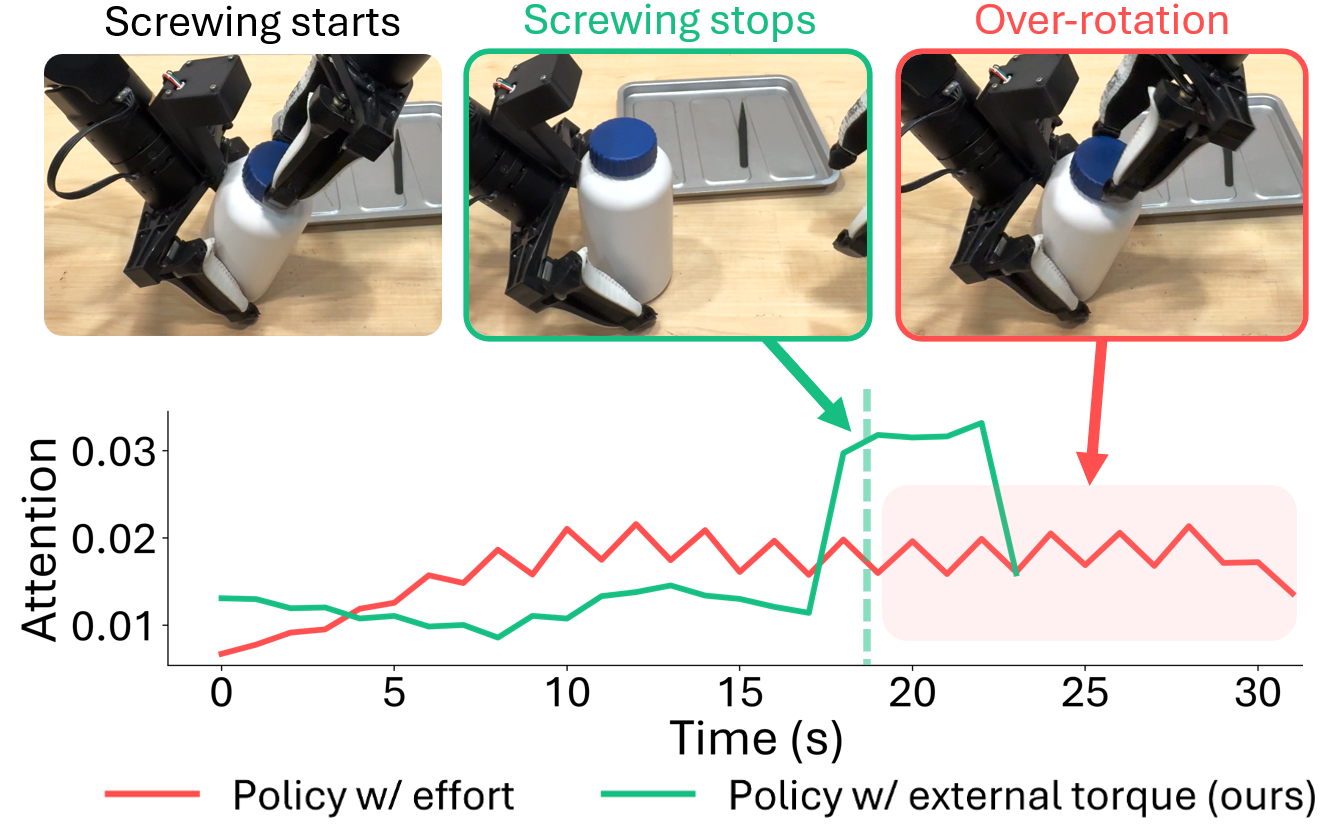}
    \caption{
    Attention values during screw-cap manipulation.
    Learned external torque produces clearer attention around key contact events than raw joint effort.
    }
    \label{fig:attention}
\end{figure}

The attention analysis in Figure~\ref{fig:attention} further supports this observation. 
To visualize how the policy uses force-related information during rollout, we compute the average cross-attention from the action tokens to the observation tokens in the first decoder layer. 
The decoder attends over both visual tokens and force-related tokens, but in Figure~\ref{fig:attention} we plot only the attention mass assigned to the force-related tokens: raw joint-current tokens for the current-conditioned policy, and learned external-torque tokens for our policy. 
This visualization therefore measures how strongly the policy attends to the force input when predicting actions.

During screw-cap manipulation, the policy conditioned on learned external torque shows a clear increase in attention around key contact events, especially when the cap becomes tightened and the screwing motion should stop. 
In contrast, the policy conditioned on raw joint current exhibits weaker and less event-aligned attention to its current tokens. 
While attention is only a diagnostic and does not by itself prove causality, this pattern suggests that learned external joint torque provides a cleaner and more task-relevant contact representation than raw joint current for force-sensitive manipulation.

\begin{table}[t]
\small
\centering
\setlength{\tabcolsep}{7pt}
\renewcommand{\arraystretch}{1.15}
\begin{tabular}{lcc}
\toprule
\textbf{Method} & \textbf{NIST Insertion $\uparrow$} & \textbf{Cap Screwing $\uparrow$} \\
\midrule
ACT & $0.220$ & $0.440$ \\
ACT w/ current & $0.431$ & $0.440$ \\
\textbf{ACT w/ $\tau_{ext}$} & $\mathbf{0.567}$ & $\mathbf{0.525}$ \\
\bottomrule
\end{tabular}
\vspace{1mm}
\caption{
Policy performance when conditioning on raw joint current versus learned external joint torque.
Learned external joint torque provides the best performance on both contact-sensitive tasks.
Higher is better. Bold indicates the best result within each task.
}
\label{tab:current_vs_external_torque}
\vspace{-5mm}
\end{table}

\subsection{Per-batch Up-sampling vs. Weighted Regression}

We further compare FIRST against a weighted-regression baseline, where the loss is reweighted according to whether each sample belongs to the free-motion, pre-contact, or contact phase.
Both approaches aim to emphasize contact-relevant data, but they do so in different ways: weighted regression modifies the optimization objective, whereas FIRST changes the training distribution by increasing the frequency of informative samples within each batch.

As shown in Table~\ref{tab:weighted_loss_sampling_lego_margin}, weighted regression either degrades performance or provides only limited improvement over the no-upsampling baseline.
In contrast, weighted sampling consistently improves LEGO assembly performance.
We observe similar qualitative trends in other contact-rich tasks.
This suggests that directly increasing the loss contribution of sparse contact-related samples can make optimization less stable, potentially causing the policy to overfit to high-weighted segments or ignore the broader trajectory context.
By contrast, per-batch up-sampling preserves the original regression objective while exposing the policy more frequently to critical pre-contact and contact transitions, leading to more stable policy learning.
\begin{table}[h]
\centering
\small
\setlength{\tabcolsep}{7pt}
\renewcommand{\arraystretch}{1.15}
\begin{tabular}{llc}
\toprule
\textbf{Phase Emphasis Mechanism} 
& \textbf{Emphasized Phase} 
& \textbf{Task Progress} \\
\midrule
None
& -- 
& 0.453 \\

\midrule

\multirow{2}{*}{Weighted Regression}
& PC
& 0.345 \\
& PC + C
& 0.478 \\

\midrule

\multirow{2}{*}{\textbf{FIRST (Ours)}}
& PC
& 0.573 \\
& PC + C
& 0.501 \\

\bottomrule
\end{tabular}
\vspace{0.5mm}
\begin{flushleft}
\end{flushleft}
\vspace{1mm}
\caption{
Comparison of phase emphasis mechanisms for LEGO assembly using ACT.
}
\label{tab:weighted_loss_sampling_lego_margin}
\end{table}

\clearpage
\end{document}